\documentclass[pdflatex,sn-mathphys-num]{sn-jnl}

\usepackage{graphicx}%
\usepackage{multirow}%
\usepackage{amsmath,amssymb,amsfonts}%
\usepackage{amsthm}%
\usepackage{mathrsfs}%
\usepackage[title]{appendix}%
\usepackage{xcolor}%
\usepackage{textcomp}%
\usepackage{manyfoot}%
\usepackage{booktabs}%
\usepackage{algorithm}%
\usepackage{algorithmicx}%
\usepackage{algpseudocode}%
\usepackage{listings}%

\usepackage{tikz}
\usetikzlibrary{shapes.geometric}
\usetikzlibrary{arrows.meta,arrows}

\usepackage{amssymb}
\usepackage{bm}
\usepackage{comment}
\usepackage{caption}
\usepackage{subcaption}
\usepackage{diagbox}
\usepackage{multirow}

\usepackage{pgfplots}
\usepgfplotslibrary{groupplots}

\pgfplotsset{compat=1.18}

\pgfplotsset{
  colormap={gainmap}{
    rgb255(0cm)=(210,115,55);
    rgb255(1cm)=(248,248,248);
    rgb255(2cm)=(70,120,185)
  }
}

\theoremstyle{thmstyleone}%
\newtheorem{theorem}{Theorem}
\newtheorem{corollary}[theorem]{Corollary}%

\theoremstyle{thmstyletwo}%

\theoremstyle{thmstylethree}%

\begin{document}

\title[LNML-based Causal Discovery with Latent Confounders: Theory and Experiments]{LNML-based Causal Discovery with Latent Confounders: Theory and Experiments}


\author*[1]{\fnm{Zhongyi} \sur{Que}}\email{que-zhongyi@g.ecc.u-tokyo.ac.jp}

\author[2]{\fnm{Shin} \sur{Matsushima}}\email{smatsus@graco.c.u-tokyo.ac.jp}

\author[1]{\fnm{Kenji} \sur{Yamanishi}}\email{yamanishi@g.ecc.u-tokyo.ac.jp}

\affil*[1]{\orgdiv{Graduate School of Information Science and Technology}, \orgname{
The University of Tokyo, Tokyo, Japan}}

\affil[2]{\orgdiv{Graduate School of Arts and Sciences}, \orgname{The University of Tokyo, Tokyo, Japan}}

\begingroup
\renewcommand{\thefootnote}{}
\footnotetext{An earlier version of this article appeared in the Proceedings of ECML PKDD 2026 under the title ``MDL Meets Latent Confounders: LNML-Based Causal Discovery.'' This article substantially extends the theoretical analysis and experimental evaluation.}
\endgroup


\abstract{Causal discovery with nonlinear mechanisms and latent confounders remains challenging. Existing methods often rely on either linear assumptions or causal sufficiency, limiting their applicability. We propose an MDL-based causal discovery framework that accommodates latent confounders and flexible nonlinear mechanisms using the luckiness normalized maximum likelihood (LNML) code-length. To complement code-length-based model selection, we introduce $\Delta$-pseudo-collinearity as a criterion for identifying potential latent-confounding candidates. We theoretically show that, in the presence of a latent confounder, the directional code-length gap $\Delta$ admits an upper bound that grows at most linearly with the sample size and decreases as the latent causal mechanisms become more similar, providing support for the proposed criterion. Based on this formulation, we develop a greedy algorithm, termed Pseudo-Collinearity Guided Causal Discovery (PCG-CD). Experiments on synthetic and real-world datasets demonstrate the ability of PCG-CD to recover directed causal relationships and identify potential latent-confounding candidates, while revealing limitations in latent-confounder discrimination.}

\keywords{Minimum description length principle, Causal discovery, Latent confounder, LNML code-length.}



\maketitle

\section{Introduction}

\subsection{Motivation}
\label{sec:motivation}
Understanding causation is a fundamental problem in science, as it enables the identification of underlying relationships among variables in complex systems.

Most existing causal discovery methods assume causal sufficiency, meaning that all relevant variables are observed and no latent confounders exist. However, confounding factors are ubiquitous in real-world systems and pose major challenges for causal estimation. When analyzing the relationship between two variables $X$ and $Y$, the observed dependence may arise from a direct causal effect, or from a latent confounder $Z$ acting as a common cause. Only a limited number of approaches relax the causal sufficiency assumption, yet failing to account for confounders can introduce substantial bias and lead to incorrect conclusions.

In addition, causal mechanisms in practical applications, such as biology, neuroscience, economics, and climate science, are rarely linear. Restricting models to linear relationships risks model misspecification and may further obscure causal structure, particularly in the presence of latent factors. Incorporating nonlinear mechanisms is therefore essential for improving identifiability and robustness in causal discovery.

\begin{figure}[ht]
  \centering
  \begin{subfigure}[t]{0.24\textwidth}
    \centering
    \begin{tikzpicture}[scale=0.2][
    \tikzstyle{every node}=[font=\normalsize]
    \draw [ color={rgb,255:red,255; green,38; blue,0} , line width=1pt ] (9.75,0.75) rectangle (17.25,-4.5);
    \draw [ line width=1pt ] (11.5,-2.5) circle (0.75cm);
    \draw [ line width=1pt ] (15.25,-2.5) circle (0.75cm);
    \draw [line width=1pt, ->, >=Stealth] (8.75,-1) -- (10.75,-2.25);
    \draw [line width=1pt, ->, >=Stealth] (8.75,-3.75) -- (10.75,-2.75);
    \draw [line width=1pt, ->, >=Stealth] (18.5,-0.5) -- (16,-2.25);
    \draw [line width=1pt, ->, >=Stealth] (18.5,-2) -- (16,-2.5);
    \draw [line width=1pt, ->, >=Stealth] (18.5,-4.25) -- (16,-2.75);
    \node [font=\normalsize] at (11.5,-2.25) {};
    \node [font=\normalsize] at (11.5,-2.25) {};
    \node [font=\normalsize] at (11.5,-2.25) {};
    \node [font=\tiny] at (15.25,-2.5) {Y};
    \node [font=\tiny] at (11.5,-2.5) {X};
    \node [font=\normalsize] at (11.5,-2.5) {};
    \draw [ color={rgb,255:red,0; green,163; blue,215}, line width=1pt, ->, >=Stealth] (12.25,-2.5) -- (14.5,-2.5);
\end{tikzpicture}
    \caption{$X$ causes $Y$.}
  \end{subfigure}
  \hfill
  \begin{subfigure}[t]{0.24\textwidth}
    \centering
    \begin{tikzpicture}[scale=0.2][
    \tikzstyle{every node}=[font=\normalsize]
    \draw [ color={rgb,255:red,255; green,38; blue,0} , line width=1pt ] (9.75,0.75) rectangle (17.25,-4.5);
    \draw [ line width=1pt ] (11.5,-2.5) circle (0.75cm);
    \draw [ line width=1pt ] (15.25,-2.5) circle (0.75cm);
    \draw [line width=1pt, ->, >=Stealth] (8.75,-1) -- (10.75,-2.25);
    \draw [line width=1pt, ->, >=Stealth] (8.75,-3.75) -- (10.75,-2.75);
    \draw [line width=1pt, ->, >=Stealth] (18.5,-0.5) -- (16,-2.25);
    \draw [line width=1pt, ->, >=Stealth] (18.5,-2) -- (16,-2.5);
    \draw [line width=1pt, ->, >=Stealth] (18.5,-4.25) -- (16,-2.75);
    \node [font=\normalsize] at (11.5,-2.25) {};
    \node [font=\normalsize] at (11.5,-2.25) {};
    \node [font=\normalsize] at (11.5,-2.25) {};
    \node [font=\tiny] at (15.25,-2.5) {Y};
    \node [font=\tiny] at (11.5,-2.5) {X};
    \node [font=\normalsize] at (11.5,-2.5) {};
    \draw [ color={rgb,255:red,0; green,163; blue,215}, line width=1pt, ->, >=Stealth] (14.5,-2.5) -- (12.25,-2.5);
\end{tikzpicture}
    \caption{$Y$ causes $X$.}
  \end{subfigure}
  \hfill
  \begin{subfigure}[t]{0.24\textwidth}
    \centering
    \begin{tikzpicture}[scale=0.2][
    \tikzstyle{every node}=[font=\normalsize]
    \draw [ color={rgb,255:red,255; green,38; blue,0} , line width=1pt ] (9.75,0.75) rectangle (17.25,-4.5);
    \draw [ line width=1pt ] (11.5,-2.5) circle (0.75cm);
    \draw [ line width=1pt ] (15.25,-2.5) circle (0.75cm);
    \draw [line width=1pt, ->, >=Stealth] (8.75,-1) -- (10.75,-2.25);
    \draw [line width=1pt, ->, >=Stealth] (8.75,-3.75) -- (10.75,-2.75);
    \draw [line width=1pt, ->, >=Stealth] (18.5,-0.5) -- (16,-2.25);
    \draw [line width=1pt, ->, >=Stealth] (18.5,-2) -- (16,-2.5);
    \draw [line width=1pt, ->, >=Stealth] (18.5,-4.25) -- (16,-2.75);
    \node [font=\normalsize] at (11.5,-2.25) {};
    \node [font=\normalsize] at (11.5,-2.25) {};
    \node [font=\normalsize] at (11.5,-2.25) {};
    \node [font=\tiny] at (15.25,-2.5) {Y};
    \node [font=\tiny] at (11.5,-2.5) {X};
    \node [font=\normalsize] at (11.5,-2.5) {};
    \draw [ color={rgb,255:red,192; green,192; blue,192} , line width=1pt ] (13.25,-0.25) circle (0.75cm);
    \node [font=\tiny, color={rgb,255:red,170; green,170; blue,170}] at (13.25,-0.25) {Z};
    \draw [ color={rgb,255:red,0; green,163; blue,215}, line width=1pt, ->, >=Stealth] (12.75,-0.75) -- (11.75,-1.75);
    \draw [ color={rgb,255:red,0; green,163; blue,215}, line width=1pt, ->, >=Stealth] (13.75,-0.75) -- (15,-1.75);
\end{tikzpicture}
    \caption{Latent confounder exists.}
  \end{subfigure}
  \hfill
  \begin{subfigure}[t]{0.24\textwidth}
    \centering
    \begin{tikzpicture}[scale=0.2][
    \tikzstyle{every node}=[font=\normalsize]
    \draw [ color={rgb,255:red,255; green,38; blue,0} , line width=1pt ] (9.75,0.75) rectangle (17.25,-4.5);
    \draw [ line width=1pt ] (11.5,-2.5) circle (0.75cm);
    \draw [ line width=1pt ] (15.25,-2.5) circle (0.75cm);
    \draw [line width=1pt, ->, >=Stealth] (8.75,-1) -- (10.75,-2.25);
    \draw [line width=1pt, ->, >=Stealth] (8.75,-3.75) -- (10.75,-2.75);
    \draw [line width=1pt, ->, >=Stealth] (18.5,-0.5) -- (16,-2.25);
    \draw [line width=1pt, ->, >=Stealth] (18.5,-2) -- (16,-2.5);
    \draw [line width=1pt, ->, >=Stealth] (18.5,-4.25) -- (16,-2.75);
    \node [font=\normalsize] at (11.5,-2.25) {};
    \node [font=\normalsize] at (11.5,-2.25) {};
    \node [font=\normalsize] at (11.5,-2.25) {};
    \node [font=\tiny] at (15.25,-2.5) {Y};
    \node [font=\tiny] at (11.5,-2.5) {X};
    \node [font=\normalsize] at (11.5,-2.5) {};
\end{tikzpicture}
    \caption{Independence.}
  \end{subfigure}
  \caption{Four possible causal relationships between two variables $X$ and $Y$.}
  \label{fig:illu_4}
\end{figure}
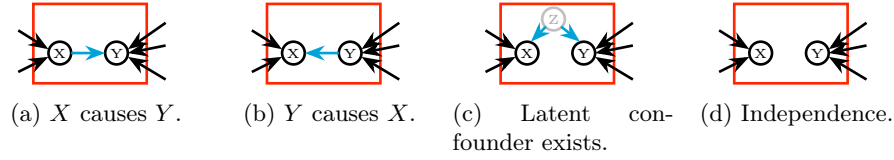

Kaltenpoth and Vreeken \cite{DK19a} suggested that an information-theoretic approach can help mitigate overfitting when estimating causal models with confounders. For example, suppose $X \not\perp\!\!\!\perp Y$ is observed. If $X$ causes $Y$ and no confounder is present, the joint distribution follows $P(X,Y) = P(X)P(Y|X)$. Alternatively, if a confounder $Z$ influences both $X$ and $Y$, the joint distribution becomes $P(X,Y,Z) = P(Z)P(X|Z)P(Y|Z)$. A straightforward method to identify a confounder is to check whether we can construct $\hat{Z}$ such that $X \perp\!\!\!\perp Y \mid Z$. A trivial solution is to set $\hat{Z} = X$, which ensures conditional independence even in the absence of a true confounder. Therefore, it is necessary to consider both the complexity of $\hat{Z}$ and its effect on $X$ and $Y$.

Motivated by this observation, we adopt the minimum description length (MDL) principle~\cite{RJ78a,GP07a,YK23a} for causal discovery. Under this framework, the causal relationship between any two variables, whether $X \rightarrow Y$, $X \leftarrow Y$, $X \leftarrow Z \rightarrow Y$, or independence, is determined by selecting the model with the shortest code-length, as illustrated in Figure~\ref{fig:illu_4}. We employ the luckiness normalized maximum likelihood (LNML) code-length~\cite{GP07a, KM17a}, which avoids explicit parameter priors and naturally incorporates sparsity through the luckiness function.

In this work, we address causal discovery with latent confounders by estimating nonlinear causal models through LNML-based model selection under the MDL principle, and theoretically characterize the resulting pseudo-collinearity criterion under general nonlinear mechanisms.

\subsection{Related Work}

Classical causal discovery methods include constraint-based and score-based approaches. Constraint-based methods such as PC \cite{PS00a} infer a causal graph by testing conditional independencies and applying orientation rules, while FCI extends PC to allow latent confounders. Score-based methods, such as GES \cite{DC02a} and its variants \cite{JR17a}, employ greedy search with penalized likelihood criteria but may converge to local optima.

To achieve identifiability beyond Markov Equivalence Classes (MECs), functional causal models (FCMs) impose assumptions on structural equations and noise distributions, including LiNGAM \cite{SS06a} with linear non-Gaussian mechanisms and ANM \cite{PH08a} with nonlinear relationships. Within the LiNGAM framework, ICA-LiNGAM \cite{SS06a} and DirectLiNGAM \cite{SS11a} recover full causal orderings under causal sufficiency, while ParceLiNGAM \cite{TT14a} and RCD \cite{TM20a} address unmeasured confounding by identifying variable pairs influenced by latent common causes and producing partially directed graphs with bidirected edges. Other extensions consider post-nonlinear assumptions \cite{KZ09a} or address latent confounding in restricted settings, including CoCa \cite{DK19a} for nonlinear pairwise relations, CLOUD \cite{MK22a} for discrete data, and NOCADILAC \cite{AK23a}, which models latent factors under a post-nonlinear assumption.

Information-theoretic approaches formulate causal discovery as a model selection problem under the MDL principle. Early methods such as SLOPE \cite{AM17a} and ORIGO \cite{KB18a} focus on bivariate or local causal direction, while GLOBE \cite{OM21a} extends MDL-based scoring to multivariate graphs under causal sufficiency. To relax the causal sufficiency assumption, latent-variable extensions based on MDL have been explored in model selection and causal discovery. For example, the decomposed normalized maximum likelihood (DNML) code-length criterion was proposed for hierarchical latent variable models \cite{KY19a}, while methods such as CDHC \cite{DK23a} introduce hidden variables when doing so reduces the overall description length, though existing results mainly focus on linear Gaussian mechanisms.

Unlike existing MDL-based approaches that primarily assume linear Gaussian mechanisms or focus on limited settings, our method performs multivariate causal graph discovery with latent confounders under nonlinear mechanisms in continuous data, using LNML code-lengths for principled model selection.

\subsection{Significance and Novelty}
\begin{itemize}
    \item \textit{LNML-based code-length for causal models with latent confounders.} 
    We introduce an LNML-based code-length formulation for estimating causal graphs in the presence of latent confounders. By adopting a tractable approximation under mild assumptions on the causal equations, the proposed framework enables efficient model selection for flexible nonlinear causal mechanisms. To the best of our knowledge, this is the first MDL-based approach to multivariate causal graph discovery with latent confounders under nonlinear mechanisms.
    
    \item \textit{Detection of potential latent confounding via pseudo-collinearity.} 
    We propose the notion of \(\Delta\)-pseudo-collinearity to detect potential latent confounding through directional code-length differences. We theoretically show that, under latent confounding, \(\Delta\) grows at most linearly with the sample size and becomes smaller as the latent causal mechanisms become more similar, providing theoretical support for the sample-size-scaled threshold used in the proposed causal discovery algorithm.
    
    \item \textit{PCG-CD: an efficient causal discovery algorithm.} 
    We develop the PCG-CD algorithm, a greedy causal discovery algorithm that explicitly handles latent confounders. Compared to GES and related score-based methods with \(O(|V|^3)\) complexity, PCG-CD reduces edge evaluation to \(O(|V|^2)\).
    
    \item \textit{Empirical validation.} Extensive experiments on synthetic and real-world datasets evaluate both $\Delta$-pseudo-collinearity and PCG-CD. The results demonstrate the usefulness of $\Delta$-pseudo-collinearity for distinguishing causal relationships under nonlinear mechanisms and the ability of PCG-CD to recover directed causal structures, while revealing the challenges of latent-confounder identification in multi-node settings.
\end{itemize}

This article extends our previous conference paper \cite{ZQ26a} in the following aspects:
\begin{itemize}
    \item \textbf{Extension of theoretical analysis to general nonlinear mechanisms.} The previous conference paper established an upper bound on the directional code-length difference $\Delta$ under linear causal mechanisms. In this article, we extend the analysis to general nonlinear mechanisms by assuming twice continuously differentiable causal functions and imposing regularity conditions on their derivatives and the associated kernel matrices. By decomposing $\Delta$ into marginal, data-fit, complexity, and normalization terms, we derive an upper bound that grows at most linearly with the sample size. We further examine quadratic mechanisms to illustrate how nonlinear terms affect the bound, particularly its dependence on latent-variable variance, and recover the linear result as a special case.
    \item \textbf{Additional evaluation of $\Delta$-pseudo-collinearity.} We extend the pairwise experiments to settings with observable parents and investigate the sensitivity of causal relationship classification to the threshold parameter $\epsilon$. These experiments provide further evidence on the applicability and limitations of the proposed criterion under different causal mechanisms and experimental settings.
    \item \textbf{Further evaluation of causal graph discovery.} We investigate the effect of graph density on PCG-CD by generating multi-node causal graphs with controlled numbers of edges. Additional evaluation metrics are introduced to examine latent-confounder recovery, directed causal relationships, and graph structure estimation.
\end{itemize}
We also correct implementation errors in the LNML code-length calculations of the previous conference version and reevaluate PCG-CD using the corrected implementation.

Code and supplementary material are available at: \url{https://github.com/ZYQue/PCG-CD}.

\section{Notation and Preliminaries}
\label{sec:CDCM}

We introduce the notation and assumptions for causal models, first under causal sufficiency and then extending to settings with latent confounders. We also summarize the MDL-based code-length formulation used throughout the paper.

\subsection{Causal Model}
\label{sec:notation}

We begin with a setting with causal sufficiency. A causal model is defined as $M=(G,F)$, where $G=(\bm{V},\bm{E})$ is a directed acyclic graph (DAG) over continuous variables $\bm{V}=\bm{X}\in\mathbb{R}^d$, and $F=\{f_i\}_{i=1}^d$ denotes the causal mechanisms. An edge $X_{(i)}\to X_{(j)}$ indicates that $X_{(i)}$ is a parent of $X_{(j)}$. 

The causal relationships between variables are described by structural equation models (SEMs) \cite{JP00a}:
\begin{equation} \nonumber
    X_{(i)} = f_i(\text{pa}_i) + \varepsilon_i, \quad \varepsilon_i \perp\!\!\!\perp \text{pa}_i,
\end{equation}
where $\text{pa}_i$ denotes the parents of $X_{(i)}$ in $G$, $f_i$ is the causal function, and $\varepsilon_i$ represents noise.

We then relax causal sufficiency by allowing latent confounders. Let $\bm{Z}$ denote unobserved variables that may jointly influence observed variables. The DAG including the confounders is then represented as $G = (\bm{V}, \bm{E})$, where $\bm{V} = (\bm{X}, \bm{Z})$. The causal relationship for variable $X_{(i)}$ becomes:
\begin{equation} \nonumber
    X_{(i)} = f_i(\text{pa}_i, \text{pa}_i^Z) + \varepsilon_i, \quad \varepsilon_i \perp\!\!\!\perp (\text{pa}_i, \text{pa}_i^Z),
\end{equation}
where $\text{pa}_i$ denotes the set of observed parents of $X^{(i)}$ in $G$, and $\text{pa}_i^Z$ denotes the set of latent parents of $X^{(i)}$.

\subsection{Minimum Description Length Principle}
According to the MDL principle, the optimal model is the one with the minimum normalized maximum likelihood (NML) code-length~\cite{YS87a}. In this work, we adopt the luckiness normalized maximum likelihood (LNML) code-length~\cite{GP07a} for causal model selection. LNML extends the original NML formulation by incorporating prior knowledge through a penalty term. This allows us to enforce sparsity in the causal DAG and to control the complexity of the causal mechanisms. In our framework, the LNML code-length is computed from the normalized maximum posterior distribution, incorporating a sparsity constraint.

The LNML distribution is defined over a model class $\mathcal{M}$, where $p(\bm{X};M)$ denotes the probability density of data $X$ under model $M$, and $g(M)$ is a penalty function that controls model complexity. It is given by
\begin{equation}
    p_{\text{LNML}} (\bm{X}) := \frac{\max_{M \in \ \mathcal{M}} p(\bm{X};M) e^{-g(M)}}{C_n}, \label{eq:LNML_prim}
\end{equation}
where
\begin{equation}
    C_n := \int \max_{M \in \mathcal{M}} p(\bm{X};M) e^{-g(M)} d \bm{X}, \nonumber
\end{equation}
also referred to as the parametric complexity, is evaluated over all possible data $\bm{X}$ of length $n$.

The LNML code-length is defined as the negative log-likelihood of \eqref{eq:LNML_prim}: 
\begin{align}
    \mathcal{L}_{\text{LNML}}(\bm{X}) &= -\log p_{\text{LNML}}(\bm{X}) \nonumber\\
    &= \underset{M \in \mathcal{M}}{\min} \{ -\log p(\bm{X};M) + g(M) \} + \log C_n. \label{eq:LNML_codelength}
\end{align}
When the penalty term $g(\cdot)$ is omitted, \eqref{eq:LNML_codelength} reduces to the original NML code-length. Computing the LNML code-length is intractable due to the parametric complexity term $C_n$, so an approximation is used. To preserve MDL properties, the approximation is constructed as an upper bound on $C_n$, whose form depends on the probability model $p(\cdot)$, the penalty function $g(\cdot)$, and the data assumptions.

\section{Luckiness Normalized Maximum Likelihood Code-length for Causal Model Discovery}
To flexibly model nonlinear causal mechanisms and enable principled model selection in the presence of latent confounders, we assume Gaussian process models and develop an LNML-based code-length for scoring causal models.

\subsection{Assumptions on the Causal Mechanism}
\label{sec:GP}

We model causal mechanisms using Gaussian processes (GPs)~\cite{CR06a}. A GP is a collection of random variables such that any finite subset follows a multivariate normal distribution, and the associated function satisfies
\[
f(x) \sim \mathcal{GP}(m(x), \kappa(x,x')),
\]
where $m(x)=\mathbb{E}[f(x)]$ is the mean function, with the expectation taken with respect to the Gaussian process prior over functions, and $\kappa(x,x')$ is the covariance (kernel) function.

Assuming GP mechanisms removes the restrictive linearity assumption between a variable and its parents. A GP can also be represented as a linear model in a (possibly infinite-dimensional) feature space: $f=\sum_i \alpha_i \phi_i$.

Given data $\{(x_i,y_i)\}_{i=1}^n$, we assume
\[
y_i = f(x_i) + \varepsilon_i,\quad \varepsilon_i \sim \mathcal{N}(0,\sigma^2),
\]
so that the outputs $\bm{y}$ follow a joint Gaussian distribution:
\[
\bm{y} \sim \mathcal{N}(\bm{m}, \bm{K} + \sigma^2 \bm{I}),
\]
where $\bm{m}=[m(x_1),\ldots,m(x_n)]^T$ and $\bm{K}_{ij}=\kappa(x_i,x_j)$.

For a given $x$, the mechanism can be estimated as:
\begin{align}
f(x) &= [\kappa(x_1,x),\ldots,\kappa(x_n,x)]^T(\bm{K}+\sigma^2\bm{I})^{-1}\bm{y} \nonumber\\
     &= \sum_{i=1}^n \alpha_i \kappa(x_i,x), \label{eq:f(x)}
\end{align}
where $\bm{\alpha}=(\bm{K}+\sigma^2\bm{I})^{-1}\bm{y}$ and $\alpha_i$ is the $i$th value of this vector.

GPs are adopted for two reasons. First, the induced reproducing kernel Hilbert space (RKHS) $\mathcal{H}$ may be infinite-dimensional, enabling a nonparametric representation of causal mechanisms and their code-lengths. Second, as shown by Kakade et al.~\cite{SK05a}, GP-based learning minimizes penalized regret, which aligns with the minimax formulation underlying NML code-length~\cite{YS87a}. Note that not all model classes admit a valid NML distribution.

We use the radial basis function (RBF) kernel
\begin{equation}
    \kappa(x,x') = \sigma_f^2 \exp\!\left(-\tfrac{1}{2l}\|x-x'\|_2^2\right), \label{eq:kernel}
\end{equation}
as it is widely used in practice.

\subsection{Luckiness Normalized Maximum Likelihood Code-length}
We first define the MDL score without confounders based on Gaussian process (GP) mechanisms. For a target variable $X_{(i)}$ with parents $\text{pa}_i(G)$ in a causal graph $G$, we fit a GP regression $f_i:\text{pa}_i(G)\to X_{(i)}$. Let $p(X_{(i)}|\text{pa}_i(G))$ denote the conditional probability of $X_{(i)}$ given its parent variables $\text{pa}_i(G)$, as induced by the Gaussian process regression model. The corresponding Bayesian score is given by~\cite{SK04a,SK05a}
\begin{align} 
    L(X_{(i)}|\text{pa}_i(G)) 
    := & \underset{f_i \in \mathcal{H}_\kappa}{\min} \big( -\log p(X_{(i)} | \text{pa}_i(G))   \nonumber\\
    & \qquad\qquad + \|f_i\|_\kappa^2 \big)+ \tfrac{1}{2} \log \det \left( \sigma^{-2} \bm{K}_{\text{pa}_i(G)} + \bm{I} \right),\label{eq:mdl_score}
\end{align}
where $\|f_i\|_\kappa^2$ penalizes the complexity of the mechanism. For GP models learned from finite data, this term can be computed as
\[
\|f_i\|_\kappa^2 = \bm{\alpha}^T \bm{K}_{\text{pa}_i(G)} \bm{\alpha},
\]
with kernel matrix $\bm{K}_{\text{pa}_i(G)}$ constructed using the kernel function defined in \eqref{eq:kernel} and vector $\bm{\alpha}$ defined in \eqref{eq:f(x)}.

The score in \eqref{eq:mdl_score} can be viewed as an upper bound on the negative log-likelihood under a Bayesian prior, assuming the model follows a Gaussian process \cite{SK04a}. As shown by Kakade et al.~\cite{SK05a}, this Bayesian prior distribution and the penalized NML density correspond to the same underlying distribution, where $\|f_i\|_\kappa^2$ acts as a luckiness function in the LNML framework~\cite{GP07a}.

The overall description length of a causal graph $G$ is defined as
\begin{align} \nonumber
    L(\bm{X}|G)
    := \sum_{i=1}^d L(X_{(i)}|\text{pa}_i(G)),
\end{align}
and the optimal causal model $\hat{M}=(\hat{G},\hat{F})$, where $\hat{G}$ denotes the estimated causal graph and $\hat{F}$ denotes the corresponding estimated causal mechanisms, is obtained by first solving
\begin{equation}
    \hat G=\arg\min_G L(X|G), \nonumber
\end{equation}
after which $\hat F$ is estimated by Gaussian process regression.

We now extend this formulation to settings with latent confounders. Consider two variables $X$ and $Y$ and a confounder $Z$. Given a causal graph $G$ with known relationships, we can infer the unknown relationship between $X$ and $Y$ by comparing the description lengths of four candidate models:
\[
X\to Y,\quad Y\to X,\quad X\leftarrow Z\rightarrow Y,\quad X\nleftrightarrow Y .
\]
The MDL score for this reduced system is
\begin{align} 
    L(X,Y|\text{pa}_X,\text{pa}_Y,Z)
    := \min \bigl\{ &L(X \rightarrow Y), L(Y \rightarrow X),\nonumber\\
    &L(X \leftarrow Z \rightarrow Y), L(X\nleftrightarrow Y) \bigr\}, \label{eq:mdl_XYZ}
\end{align}
where
\begin{align} 
    &L(X \rightarrow Y) := L(Y|\text{pa}_Y,X) + L(X|\text{pa}_X), \nonumber\\
    &L(Y \rightarrow X) := L(X|\text{pa}_X,Y) + L(Y|\text{pa}_Y), \nonumber \\
    &L(X \leftarrow Z \rightarrow Y) := L(X|\text{pa}_X,Z) + L(Y|\text{pa}_Y,Z), \nonumber \\
    &L(X\nleftrightarrow Y) := L(X|\text{pa}_X) + L(Y|\text{pa}_Y). \label{eq:4types_origin}
\end{align}
Since the MDL score in \eqref{eq:mdl_score} penalizes the complexity of $f$, setting $\hat{Z} = X$ would maximize the likelihood but would fail to minimize the description length. This approach avoids the overfitting issue discussed by Kaltenpoth and Vreeken \cite{DK19a}.

Finally, for a causal model $M=(G,F)$ with observed variables and latent confounders, the resulting computable LNML objective is
\begin{align} 
    \bar{\mathcal{L}}_{\text{LNML}}(\bm{X})
    := & 
    \sum_{i=1}^d \biggl\{ 
    -\log p(X_{(i)}|\text{pa}_i(G)) \nonumber \\
    & \qquad \qquad \qquad + \|f_i\|_\kappa^2 + \tfrac{1}{2} \log \det \left( \sigma^{-2} \bm{K}_{\text{pa}_i(G)} + \bm{I} \right) \biggr\}. \label{eq:computable_LNML}
\end{align}
The optimal graph is obtained by minimizing
$\bar{\mathcal{L}}_{\text{LNML}}(\bm{X}|G)$ over $G$.

\section{Algorithm for Solving a Causal Model with Latent Confounders}
While LNML code-length can be defined with latent confounders, its direct computation is challenging due to reliance on sampling over unobservable variables. To address this issue, we introduce a computable approximation of the latent-confounder-aware code-length that removes sampling dependence. Based on this approximation, we propose the notion of \emph{pseudo-collinearity} to theoretically characterize directional ambiguity and detect potential latent-confounding signals. These ideas lead to a novel causal discovery algorithm, \textbf{Pseudo-Collinearity Guided Causal Discovery (PCG-CD)}, which departs from conventional score-based methods by explicitly incorporating the detection of potential latent-confounding relations into model selection.

\subsection{Reformulation of Code-length} \label{sec:reformulation_codelength}
In this work, we assume the presence of latent confounders, under which statistical dependencies may need to be jointly explained by observable variables and unobservable variables (latent confounders). To simplify the exposition, we henceforth focus on the residual components of variables when considering a pair of nodes \(A\) and \(B\).

The residual component of a node is defined as the original variable minus the part explained by its known parents in the causal graph \(G\). We denote the residual component of \(A\) by \(A_{\text{res}}\). Under this formulation, the causal relationship between \(A\) and \(B\) can be identified by examining \(A_{\text{res}}\) and \(B_{\text{res}}\), without reference to other observed nodes in \(G\).

Under this setting, we consider a reduced system consisting of $A_{\text{res}}$, $B_{\text{res}}$, and an unobserved variable $Z$. The four candidate models differ only in the assumed causal structure among these variables. We assume that the latent node $Z$ is present throughout the reduced system, but its realization is unobserved. Each candidate model specifies a different causal explanation for the pair $(A_{\text{res}},B_{\text{res}})$, and only the confounded model uses $Z$ as their common cause. Since the same latent variable $Z$ is present in all candidates, its code-length $L(Z)$ is identical across them and does not affect the model comparison. The corresponding code-lengths in \eqref{eq:4types_origin} can be rewritten as follows:
\begin{align}
    & L(A_{\text{res}}|\text{pa}_{A_{\text{res}}}=\emptyset)+L(B_{\text{res}}|\text{pa}_{B_{\text{res}}}=A_{\text{res}}) + L(Z),\nonumber\\
    & L(A_{\text{res}}|\text{pa}_{A_{\text{res}}}=B_{\text{res}})+L(B_{\text{res}}|\text{pa}_{B_{\text{res}}}=\emptyset) + L(Z), \nonumber\\
    & L(A_{\text{res}}|\text{pa}_{A_{\text{res}}}=Z)+L(B_{\text{res}}|\text{pa}_{B_{\text{res}}}=Z) + L(Z),\nonumber\\
    &L(A_{\text{res}}|\text{pa}_{A_{\text{res}}}=\emptyset)+L(B_{\text{res}}|\text{pa}_{B_{\text{res}}}=\emptyset) + L(Z). \nonumber
\end{align}
Since $L(Z)$ can be treated as constant, we can ignore it when comparing the code-lengths of the four cases. 

We can also reformulate the code-lengths of the four cases as follows:
\begin{align}
    &L(A_{\text{res}} \rightarrow B_{\text{res}}) = L(A_{\text{res}}|\text{pa}_{A_{\text{res}}}=\emptyset)+L(B_{\text{res}}|\text{pa}_{B_{\text{res}}}=A_{\text{res}}) ,\nonumber\\
    &L(A_{\text{res}} \leftarrow B_{\text{res}}) = L(A_{\text{res}}|\text{pa}_{A_{\text{res}}}=B_{\text{res}})+L(B_{\text{res}}|\text{pa}_{B_{\text{res}}}=\emptyset) ,\nonumber\\
    &L(A_{\text{res}} \leftrightarrow B_{\text{res}}) = L(A_{\text{res}}|\text{pa}_{A_{\text{res}}}=B_{\text{res}})+L(B_{\text{res}}|\text{pa}_{B_{\text{res}}}=A_{\text{res}}) ,\nonumber\\
    &L(A_{\text{res}} \nleftrightarrow B_{\text{res}}) = L(A_{\text{res}}|\text{pa}_{A_{\text{res}}}=\emptyset)+L(B_{\text{res}}|\text{pa}_{B_{\text{res}}}=\emptyset) \nonumber,
\end{align}%
where $A_{\text{res}} \leftrightarrow B_{\text{res}}$ denotes the case where a latent confounder causes both $A$ and $B$, and $A_{\text{res}} \nleftrightarrow B_{\text{res}}$ denotes the case where no causal relationship exists between $A$ and $B$. The code-length of the case with latent confounder $L(A_{\text{res}} \leftrightarrow B_{\text{res}})$ serves as an approximation to $L(A_{\text{res}}|\text{pa}_{A_{\text{res}}}=Z)+L(B_{\text{res}}|\text{pa}_{B_{\text{res}}}=Z)$ by replacing $Z$ with $A_{\text{res}}$ and $B_{\text{res}}$. This approximation is heuristic in nature. Intuitively, one may expect $L(A_{\text{res}}|\text{pa}_{A_{\text{res}}}=B_{\text{res}})$ to approach $L(A_{\text{res}}|\text{pa}_{A_{\text{res}}}=Z)$ when \(B = f(Z) + \varepsilon_B\), provided that the function \(f(\cdot)\) is simple and the noise term \(\varepsilon_B\) is small (e.g., \(B = Z + \mathcal{N}(0, 0.0001)\)). 

Since the quality of this approximation may vary with the complexity of the causal mechanisms, we do not use it to determine causal relationship types under the MDL principle. Instead, it is introduced solely to make the overall code-length of causal graphs with latent confounders computable, eliminating the need for explicit sampling of latent confounder values. This intuition extends naturally to the original nodes \(A\) and \(B\) in the causal graph \(G\).

\subsection{Pseudo-collinearity} \label{sec:psedo-collinearity}
After accounting for observable influences, if residuals of two nodes remain similarly dependent in both directions, we regard this as a potential signal of latent confounding. We call this property pseudo-collinearity.

Pseudo-collinearity holds if:
\begin{align}
    & L(A_{\text{res}}|\text{pa}_{A_{\text{res}}}=B_{\text{res}})+L(B_{\text{res}}|\text{pa}_{B_{\text{res}}}=\emptyset)\nonumber\\
    & \qquad \qquad \approx L(A_{\text{res}}|\text{pa}_{A_{\text{res}}}=\emptyset)+L(B_{\text{res}}|\text{pa}_{B_{\text{res}}}=A_{\text{res}}). \label{eq:pcl}
\end{align}
If the difference between the left-hand side and right-hand side of \eqref{eq:pcl} is large, meaning that the code-length of either \(A\to B\) or \(A\leftarrow B\) is significantly smaller, the MDL criterion provides stronger support for a unidirectional causal explanation. More specifically, we say that a pair of variables $A$ and $B$ exhibits \textbf{$\Delta$-pseudo-collinearity} if
\begin{align}
    \Delta &= |L(A_{\text{res}}|\text{pa}_{A_{\text{res}}}=\emptyset)+L(B_{\text{res}}|\text{pa}_{B_{\text{res}}}=A_{\text{res}})\label{eq:dpcl} \\ 
    & \qquad \qquad - \Big( L(A_{\text{res}}|\text{pa}_{A_{\text{res}}}=B_{\text{res}})+L(B_{\text{res}}|\text{pa}_{B_{\text{res}}}=\emptyset) \Big)|
    \leq n\epsilon, \nonumber
\end{align}
where $n$ denotes the sample size and $\epsilon$ is a predefined threshold. $\Delta$ captures the discrepancy between alternative explanatory directions, and we regard $A$ and $B$ as $\Delta$-pseudo-collinear whenever this discrepancy falls below the threshold $n\epsilon$.

\subsection{A Theoretical Evaluation on Pseudo-collinearity}

The $\Delta$ in \eqref{eq:dpcl} should be small when pseudo-collinearity exists between nodes $A$ and $B$. We begin by considering general nonlinear mechanisms between a common latent confounder $Z$ and the observed nodes $A$ and $B$, and then examine quadratic and linear cases as more specific settings.

\begin{theorem}[$\Delta$-pseudo-collinearity of General Nonlinear Case]\label{the:general_nonlinear}
Let nodes $A$ and $B$ be influenced by a common latent confounder $Z$, where $A$ and $B$ have no additional parents and
\[
A = g_A(Z) + \epsilon_A, \qquad
B = g_B(Z) + \epsilon_B.
\]
Assume that $g_A$ and $g_B$ are twice continuously differentiable and locally invertible on the effective support of $Z$, with their first derivatives bounded away from zero and infinity and their second derivatives bounded. Assume additionally that the spectra of the kernel matrices associated with \(A\) and \(B\) are comparable. Then
\[
\Delta = O( n C(g_A,g_B,\sigma_Z,\sigma_A,\sigma_B)),
\]
where $n$ denotes the sample size, $\sigma_Z$ denotes the noise level of the latent confounder, and $\sigma_A,\sigma_B$ denote the noise levels of $A$ and $B$, respectively.
Moreover, $C(g_A,g_B,\sigma_Z,\sigma_A,\sigma_B)$ decreases as the mechanisms $g_A$ and $g_B$ become more similar, while its dependence on the latent-confounder noise level $\sigma_Z$ is not necessarily monotonic.
\end{theorem}

To prove Theorem~\ref{the:general_nonlinear}, we first express the two directional relations through local inverse mappings and then decompose the LNML code-length difference into the marginal and GP data-fit terms, the complexity term, and the normalization term. Each component is shown to be bounded by a constant times $n$, and the corresponding constants decrease as the two mechanisms induce more similar marginal distributions, directional regressions, functional complexities, and kernel spectra. The complete proof is given in Appendix~\ref{apd:general}. 

Theorem~\ref{the:general_nonlinear} establishes that, under general nonlinear mechanisms, the pseudo-collinearity gap $\Delta$ grows at most linearly with the sample size. This linear scaling motivates the thresholding rule $\Delta\leq n\epsilon$ used in the algorithm.

\paragraph{Quadratic specialization.} The general result above does not give an explicit form for the dependence of $C$ on the mechanism parameters. We therefore consider quadratic mechanisms as a concrete specialization, for which the latent-induced moments and local conditional variances can be written explicitly. Let
\[
g_A(z)=\alpha_1+\alpha_2 z+\alpha_3 z^2,
\qquad
g_B(z)=\beta_1+\beta_2 z+\beta_3 z^2.
\]
In this case, the marginal moments and the local conditional variances appearing in the bound can be written explicitly. In particular, the latent contribution to the marginal variances contains both $\sigma_Z^2$ and $\sigma_Z^4$ terms, while the conditional variances depend on the local derivative ratios of $g_A$ and $g_B$. These expressions make the mechanism similarity more explicit: when the corresponding linear and quadratic coefficients become closer, the induced marginal distributions and local conditional regressions also become more similar, which reduces the corresponding contribution to $C$.

At the same time, the quadratic form reveals an important difference from the linear case. Increasing $\sigma_Z^2$ strengthens the shared latent variation, but also amplifies the quadratic contribution through the $\sigma_Z^4$ terms and changes the local nonlinear mappings between $A$ and $B$. Therefore, the dependence of $C$ on $\sigma_Z^2$ is not necessarily monotonic. The proof follows the decomposition used in Theorem~\ref{the:general_nonlinear}, but explicitly evaluates the quadratic marginal moments and local conditional variances to expose these additional terms. A detailed proof is provided in Appendix~\ref{apd:quadratic}.

\paragraph{Linear case.} The linear case provides a further specialization of Theorem~\ref{the:general_nonlinear}. Unlike the general nonlinear and quadratic cases, it admits a stronger characterization of the effect of the latent-variable noise level.

\begin{corollary}[$\Delta$-pseudo-collinearity of Linear Case]\label{cor:linear}
Let nodes $A$ and $B$ be influenced by a common latent confounder $Z$, where $A$ and $B$ have no additional parents and
\[
A=\alpha_1+\alpha_2 Z+\epsilon_A,\qquad
B=\beta_1+\beta_2 Z+\epsilon_B.
\]
Then
\[
\Delta=O\!\left(nC\!\left(|\alpha_2-\beta_2|,\sigma_Z,\sigma_A,\sigma_B\right)\right),
\]
where $n$ denotes the sample size, $|\alpha_2-\beta_2|$ denotes the difference between the coefficients of the two mechanisms $Z\rightarrow A$ and $Z\rightarrow B$, $\sigma_Z$ denotes the noise level of the latent confounder, and $\sigma_A,\sigma_B$ denote the noise levels of $A$ and $B$, respectively. Moreover, $C(|\alpha_2-\beta_2|,\sigma_Z,\sigma_A,\sigma_B)$ decreases as the noise level $\sigma_Z$ increases and as $|\alpha_2-\beta_2|$ decreases, while the effect of $\sigma_A$ and $\sigma_B$ depends on their interaction with the mechanism coefficients and other noise terms.
\end{corollary}

The proof of Corollary~\ref{cor:linear} rewrites the latent linear model as an induced linear dependence between $A$ and $B$ and directly compares the corresponding directional LNML code-lengths. In this case, the absence of higher-order terms yields the stronger result that the bound decreases as the latent-variable noise level increases. The complete derivation is given in Appendix~\ref{apd:linear}.

Taken together, these results provide a theoretical motivation for using a sample-size-scaled threshold for $\Delta$. When $A$ and $B$ are generated by a common latent confounder through similar mechanisms, the two directional code-lengths become comparable and their difference remains bounded by a quantity that grows at most linearly with $n$. In contrast, a directed causal relation does not in general possess this symmetric latent-generation structure, and there is no corresponding reason for the two directional code-lengths to remain close. We therefore use the criterion $\Delta\leq n\epsilon$ in \eqref{eq:dpcl} as a heuristic signal of $\Delta$-pseudo-collinearity and potential latent confounding, while larger values of $\Delta$ tend to favor a directed causal explanation.

\subsection{The PCG-CD Algorithm} \label{sec:DAG_algorithm}
Based on our earlier formulation of code-lengths for the four types of causal relationships and the notion of pseudo-collinearity, we propose a novel greedy causal discovery algorithm, termed Pseudo-Collinearity Guided Causal Discovery (PCG-CD), which differs fundamentally from existing score-based methods. In contrast to GES-based approaches, which have a computational complexity of \(O(|V|^3)\) for edge evaluation, the proposed method operates with a reduced complexity of \(O(|V|^2)\). Our method follows a deletion-based strategy that starts from a complete graph and progressively removes edges according to the MDL principle. An edge is removed only when its contribution to explaining the causal structure or reducing the overall code-length is negligible, and removed edges are never reintroduced. As a result, the overall code-length decreases monotonically throughout the procedure.

\begin{algorithm}[H]
\caption{PCG-CD}
\label{alg:pcgcd}
\textbf{Input}: Dataset $\{X_{(1)},\dots,X_{(d)}\}$ \\
\textbf{Parameter}: Threshold $n\epsilon$, maximum iterations $T_{\max}$ \\
\textbf{Output}: Causal graph $G$
\begin{algorithmic}
    \State Initialize graph $G$ with bidirected edge $X_{(i)} \leftrightarrow X_{(j)}$ for all $i \neq j$.
    \State Initialize $t \leftarrow 0.$
    \State \textbf{Stage 1: Skeleton Construction}
    \While{$G$ is still changing and $t<T_{\max}$}
        \ForAll{adjacent variable pairs $(X_{(i)}, X_{(j)})$}
            \If{$L(X_{(i)} \nleftrightarrow X_{(j)}) < \min\{L(X_{(i)} \rightarrow X_{(j)}),  L(X_{(j)} \rightarrow X_{(i)})\}$}
                \State Replace $X_{(i)} \leftrightarrow X_{(j)}$ by $X_{(i)} \nleftrightarrow X_{(j)}$.
            \ElsIf{$|L(X_{(i)} \rightarrow  X_{(j)}) - L(X_{(j)} \rightarrow X_{(i)})| > n\epsilon$}
                \State Keep the direction with a smaller code-length.
            \Else
                \State Keep $X_{(i)} \leftrightarrow X_{(j)}$.
            \EndIf
        \EndFor
        \State $t \leftarrow t+1.$
    \EndWhile
    
    \State \textbf{Stage 2: Cycle Removal}
    \While{$G$ contains a directed cycle}
        \ForAll{unidirectional edges $\rightarrow$ in a cycle}
            \State Compute the loss for deleting or replacing $\rightarrow$ with $\leftrightarrow$.
        \EndFor
        \State Modify the edge with the minimum loss.
    \EndWhile
    
    \State \textbf{return} $G$.
\end{algorithmic}
\end{algorithm}

PCG-CD proceeds in two stages and is designed to account for latent confounders. It starts from a fully connected graph where all variable pairs are linked by bidirected edges, representing maximal uncertainty and allowing for unobserved common causes.

In the first stage, the algorithm iteratively refines the graph by comparing LNML-based code-lengths under edge configurations for each variable pair. Edges are removed when no direct relation yields a shorter code-length, oriented when one direction is clearly favored, or kept bidirected when the code-length difference is below a threshold. The latter case reflects pseudo-collinearity, indicating that neither causal direction is strongly supported and flagging the pair as a potential latent-confounding candidate. This stage produces a partially oriented graph that encodes potential latent confounding. In the second stage, the algorithm enforces acyclicity by detecting directed cycles and modifying the edge whose adjustment causes the smallest increase in code-length, either by removal or by converting it to a bidirected edge. The final output is a directed acyclic causal graph balancing model simplicity and goodness of fit. Bidirected edges representing potential latent-confounding relations are ignored when detecting directed cycles.

In this greedy procedure, the causal mechanisms $F$ are re-estimated at each iteration by fitting Gaussian process regressions according to the current parent sets in $G$. Through the iterative refinements in the first stage and the acyclicity adjustments in the second stage, the algorithm progressively reduces the overall code-length and returns the final causal model $\hat{M}=(\hat{G},\hat{F})$.

Note that the time complexity of edge checking in this algorithm is \(O(|V|^2)\). To capture nonlinear mechanisms, we adopt a Gaussian Process assumption in the code-length definition, making GP regression the main computational cost and increasing runtime with sample size. Nevertheless, the PCG-CD edge-searching framework remains advantageous if alternative code-length definitions are adopted in other scenarios.

\section{Experiments}
We conduct four experiments on synthetic data to evaluate the performance of $\delta$-pseudo-collinearity and PCD-CD algorithm. The first two experiments examine the $\Delta$-pseudo-collinearity criterion in pairwise settings without and with observable parents. The third experiment evaluates PCG-CD on multi-node causal graphs, while the fourth investigates the effect of graph density on causal structure recovery.

For all experiments, each observed variable is standardized independently using z-score normalization before computing the LNML code-length. Gaussian process regression is performed by using the \texttt{scikit-learn} implementation. We use $\mathrm{WhiteKernel}(1.0)$ for variables without parents and $\mathrm{RBF}(1.0)+\mathrm{WhiteKernel}(1.0)$ for variables with parents. All kernel hyperparameters are initialized by default values, and optimized during training. Unless explicitly stated otherwise, the threshold parameter is fixed at $\epsilon=0.5$ across the main experiments to avoid setting-specific threshold tuning. The maximum number of iteration in the first stage of PCG-CD is set to 20.

\subsection{Performance Metrics}

For pairwise experiments, we use the area under the ROC curve (AUC) of the code-length gap $\Delta$ to evaluate whether $\Delta$-pseudo-collinearity distinguishes directly connected pairs from pairs without a direct causal relationship. For the experiment with observable parents, we additionally report three accuracy measures. \textbf{Acc1} is an exploratory measure obtained by directly selecting the minimum among the four candidate code-lengths. Because the code-length for
the latent-confounder case is only approximated, this four-way comparison is not used in the proposed algorithm; Acc1 is reported only to assess how well the approximation behaves in direct model selection. It also serves as a baseline for quantifying the improvement brought by the $\Delta$-based criterion in Acc2. \textbf{Acc2} denotes the accuracy after applying the threshold $n\epsilon$ to distinguish among the four possible relationships, $A\rightarrow B$, $A\leftarrow B$, $A\leftrightarrow B$, and $A\not\leftrightarrow B$. \textbf{Acc3} uses the same decision rule but treats $A\leftrightarrow B$ and $A\not\leftrightarrow B$ as a single class corresponding to the absence of a direct causal relationship.

For multi-node causal graph discovery, although Structural Hamming Distance (SHD) \cite{DC02a} and Structural Intervention Distance (SID) \cite{JP15a} are widely used metrics for evaluating causal discovery methods whose outputs are fully specified DAGs, methods designed to handle latent confounders (e.g., FCI) typically output Partial Ancestral Graphs (PAGs). In such graphs, both edge orientations and interventional effects are generally not uniquely identifiable. Consequently, SHD and SID are not well suited for evaluating these methods. We therefore adopt structure-aware and confounder-aware metrics that more accurately reflect the information encoded in causal graphs containing both directed causal edges and latent confounders.

We use \textbf{Z-hit}, \textbf{E-hit}, \textbf{Sim1}, and \textbf{Sim2} as the primary evaluation metrics in multi-node settings. Z-hit measures the proportion of true latent-confounded pairs $A\leftrightarrow B$ that are correctly recovered. E-hit measures the proportion of true directed relationships that are correctly identified. Sim1 quantifies the similarity between the ground-truth causal graph and the estimated graph in terms of adjacency matrices, treating $A\leftrightarrow B$ as the absence of an edge between $A$ and $B$. In contrast, Sim2 treats $A\leftrightarrow B$ as the presence of both $A\rightarrow B$ and $A\leftarrow B$.

For the graph-density experiment, we additionally consider metrics that distinguish recovery of latent-confounded pairs, direct causal edges, and overall graph structure. Details are explained in the later corresponding subsection.

\subsection{Synthetic Data}\label{sec:synthetic}

\subsubsection{Validation of $\Delta$-Pseudo-Collinearity}\label{sec:exp_delta}

\paragraph{Without observable parents.}

We first conduct experiments on pairwise nodes to examine whether $\Delta$-pseudo-collinearity can distinguish direct causal relationships from latent-confounded relationships. We assume that nodes $A$ and $B$ are dependent and have no additional observable parents. Without loss of generality, we randomly generate data under two causal relationships, $A\rightarrow B$ and $A\leftrightarrow B$, each occurring with probability of 50\%.

We consider both linear and nonlinear causal mechanisms. In the linear case, nodes without parents are generated according to
\begin{equation}
x=U(-5,5)+N(0,\sigma^2),
\end{equation}
where $U(\cdot)$ denotes the uniform distribution and $N(0,\sigma^2)$ represents additive Gaussian noise. Nodes with parents $\mathrm{pa}_x$ follow
\begin{equation}
x=\alpha_1+\alpha_2\mathrm{pa}_x+N(0,\sigma^2),
\end{equation}
where $\alpha_1$ and $\alpha_2$ are independently sampled from $U(-10,10)$.

In the nonlinear case, we adopt a cubic polynomial mechanism. Nodes without parents follow
\begin{equation}
x=U(-3,3)+N(0,\sigma^2),
\end{equation}
while nodes with parents are generated according to
\begin{equation}
x=\beta_1+\beta_2\mathrm{pa}_x+\beta_3\mathrm{pa}_x^2+\beta_4\mathrm{pa}_x^3+N(0,\sigma^2),
\end{equation}
where $\beta_1,\beta_2,\beta_3,\beta_4$ are independently sampled from $U(-2,2)$.

We consider sample sizes $n\in\{200,500,1000\}$ and noise levels $\sigma\in\{0.2,0.5,1.0\}$. For each setting, 20 independent datasets are generated. The observed variables are standardized before computing the LNML code-lengths. We evaluate the absolute difference $\Delta$ between the two directional code-lengths using AUC. A smaller $\Delta$ indicates that the two causal directions have similar code-lengths and therefore provide weak evidence for distinguishing the direction of causation.

\begin{table}[t]
\caption{AUC scores of the code-length gap $\Delta$ for distinguishing direct causal relationships from latent-confounded relationships under different noise levels $\sigma$ and sample sizes.}
\label{tab:exp1_auc}
\centering
\begin{tabular}{ccccccc}
\toprule
\multirow{2}{*}{Noise level $\sigma$} & \multicolumn{3}{c}{Linear cases} & \multicolumn{3}{c}{Nonlinear cases} \\
\cmidrule(lr){2-4}\cmidrule(lr){5-7}
& 200 & 500 & 1000 & 200 & 500 & 1000 \\
\midrule
0.2 & 0.5104 & 0.5455 & 0.4533 & 1.0000 & 1.0000 & 1.0000 \\
0.5 & 0.3956 & 0.2727 & 0.5104 & 1.0000 & 0.9394 & 0.9670 \\
1.0 & 0.4271 & 0.3750 & 0.6566 & 0.9780 & 0.9798 & 1.0000 \\
\botrule
\end{tabular}
\end{table}

Table~\ref{tab:exp1_auc} presents the AUC results under different sample sizes and noise levels. In the nonlinear cases, the AUC ranges from $0.9394$ to $1.0000$, indicating that $\Delta$ provides strong discrimination between direct and latent-confounded relationships under the considered settings. In contrast, the linear cases show substantially lower AUC values, ranging from $0.2727$ to $0.6566$. These results indicate that $\Delta$-pseudo-collinearity is less effective in distinguishing the two relationship types under the standardized linear mechanisms without observable parents. There is no consistent monotonic dependence between sample size and noise level. The effects of data standardization and kernel specifications are further examined in the Appendix \ref{app:pairwise_standardization}.

\paragraph{With observable parents.}

We next extend the pairwise experiment to a setting with observable parents. The data-generating mechanisms and coefficient distributions are the same as those used in the previous experiment. For the two target nodes $A$ and $B$, one node has one observable parent and the other has two observable parents. No causal relationships are introduced between their respective parent sets. The relationship between $A$ and $B$ is randomly selected from four possible cases: $A\rightarrow B$, $A\leftarrow B$, $A\leftrightarrow B$, and no direct relation, each with probability $0.25$. This setting allows us to examine whether $\Delta$ remains informative after additional observable causal influences have been introduced.

We consider the same sample sizes and noise levels as in the previous experiment, with 20 independent datasets for each setting. All results are obtained using the fixed threshold $\epsilon=0.5$.

\begin{table*}[t]
\caption{AUC, Acc1, Acc2, and Acc3 results for pairwise causal discovery with observable parents under different noise levels $\sigma$ and sample sizes.}
\label{tab:exp2_pairwise}
\centering
\begin{tabular*}{\textwidth}{@{\extracolsep{\fill}}cccccccccc}
\toprule
& & \multicolumn{4}{c}{Linear cases} & \multicolumn{4}{c}{Nonlinear cases} \\
\cmidrule(lr){3-6}\cmidrule(lr){7-10}
$n$ & $\sigma$ & AUC & Acc1 & Acc2 & Acc3 & AUC & Acc1 & Acc2 & Acc3 \\
\midrule
\multirow{3}{*}{200}
& 0.2 & 0.9798 & 0.80 & 0.90 & 0.90 & 1.0000 & 0.65 & 1.00 & 1.00 \\
& 0.5 & 0.9394 & 0.80 & 0.75 & 0.80 & 1.0000 & 0.55 & 1.00 & 1.00 \\
& 1.0 & 1.0000 & 0.90 & 0.90 & 0.90 & 1.0000 & 0.60 & 1.00 & 1.00 \\
\midrule
\multirow{3}{*}{500}
& 0.2 & 1.0000 & 1.00 & 1.00 & 1.00 & 1.0000 & 0.45 & 0.85 & 0.95 \\
& 0.5 & 0.9167 & 0.90 & 0.80 & 0.80 & 1.0000 & 0.65 & 0.75 & 0.75 \\
& 1.0 & 0.9899 & 0.85 & 0.95 & 0.95 & 1.0000 & 0.50 & 1.00 & 1.00 \\
\midrule
\multirow{3}{*}{1000}
& 0.2 & 0.9451 & 0.70 & 0.90 & 0.90 & 1.0000 & 0.40 & 1.00 & 1.00 \\
& 0.5 & 0.9899 & 0.85 & 0.95 & 0.95 & 1.0000 & 0.40 & 0.80 & 0.95 \\
& 1.0 & 0.9271 & 0.90 & 0.85 & 0.85 & 1.0000 & 0.60 & 0.90 & 0.90 \\
\botrule
\end{tabular*}
\end{table*}

Table~\ref{tab:exp2_pairwise} reports the results for pairwise causal discovery with observable parents. The linear cases achieve AUC values between $0.9167$ and $1.0000$, while the nonlinear cases achieve an AUC of $1.0000$ for all combinations of sample size and noise level. Compared with the experiment without observable parents, $\Delta$ provides substantially stronger discrimination in the linear cases. These results suggest that $\Delta$-pseudo-collinearity remains informative when additional observable-parent effects are present.

The accuracy results further illustrate the usefulness of the $\Delta$-based decision rule. In the nonlinear cases, Acc1 ranges from $0.40$ to $0.65$, whereas Acc2 and Acc3 range from $0.75$ to $1.00$. Thus, directly selecting the minimum among the four candidate code-lengths is less effective than applying the $\Delta$ threshold. In the linear cases, Acc1, Acc2, and Acc3 are generally high, although thresholding does not improve accuracy in every setting. Overall, these results indicate that the $\Delta$-based decision rule provides useful information beyond direct comparison of the approximate code-lengths, particularly under nonlinear mechanisms. The sensitivity of Acc2 and Acc3 to the choice of $\epsilon$ is examined in the Appendix \ref{app:exp2_threshold}.

\subsubsection{Multi-node Causal Graph Discovery}

\paragraph{Causal graph recovery.}

For datasets with multiple variables, we generate random directed acyclic graphs (DAGs) using the Erd\H{o}s--R\'enyi model with an edge probability of $0.25$. Each graph contains six nodes, of which five are observed and one is latent. One node with at least two children is randomly selected as the latent confounder. Nonlinear causal mechanisms based on cubic polynomials are used, consistent with the previous experiments. We consider sample sizes $n\in\{200,500,1000\}$, with 20 datasets for each sample size. The latent variable is excluded from causal graph estimation.

We compare PCG-CD with FCI, a constraint-based causal discovery algorithm that allows latent confounders \cite{PS20a}. FCI is implemented using \texttt{causallearn} with default FCI parameters, including the Fisher-Z conditional independence test. Since FCI produces a Partial Ancestral Graph (PAG) containing partially oriented edges, we consider two evaluation protocols. FCI evaluates only fully oriented edges ($\rightarrow$ and $\leftrightarrow$), while FCI (all) additionally interprets $\circ\!\!\rightarrow$ as $\rightarrow$ and $\circ\!-\!\circ$ as $\leftrightarrow$ for the purpose of computing evaluation metrics. For comparison, we retain the FCI and FCI (all) results from the previous experiment. For PCG-CD, the observed variables are standardized before estimation.

\begin{table}[t]
\caption{Performance comparison under different sample sizes in terms of latent-confounder recovery, directed edge recovery, and graph similarity metrics.}
\label{tab:exp3_multinode}
\centering
\begin{tabular}{clcccc}
\toprule
Size & Method & Z-hit & E-hit & Sim1 & Sim2 \\
\midrule
\multirow{3}{*}{200}
& PCG-CD & 0.50 & \textbf{0.77} & \textbf{0.92} & 0.84 \\
& FCI & 0.15 & 0.11 & 0.90 & 0.84 \\
& FCI (all) & \textbf{0.60} & 0.34 & 0.91 & \textbf{0.89} \\
\midrule
\multirow{3}{*}{500}
& PCG-CD & 0.35 & \textbf{0.73} & \textbf{0.92} & 0.86 \\
& FCI & 0.00 & 0.15 & \textbf{0.92} & 0.84 \\
& FCI (all) & \textbf{0.40} & 0.33 & 0.91 & \textbf{0.87} \\
\midrule
\multirow{3}{*}{1000}
& PCG-CD & 0.40 & \textbf{0.73} & \textbf{0.91} & \textbf{0.88} \\
& FCI & 0.05 & 0.07 & 0.90 & 0.82 \\
& FCI (all) & \textbf{0.45} & 0.39 & 0.89 & 0.85 \\
\botrule
\end{tabular}
\end{table}

Table~\ref{tab:exp3_multinode} presents the performance of PCG-CD and the two FCI evaluation protocols. PCG-CD achieves Z-hit values of $0.50$, $0.35$, and $0.40$ for sample sizes 200, 500, and 1000, respectively. These results indicate that recovery of the true latent-confounded pair remains challenging in multi-node settings. PCG-CD achieves higher Z-hit than standard FCI across all three sample sizes, but its Z-hit values are lower than those of FCI (all). This difference reflects the evaluation of partially oriented edges under the FCI (all) protocol and should not be interpreted as equivalent confidence in the recovered latent-confounded relations.

For directed causal relationships, PCG-CD achieves E-hit values ranging from $0.73$ to $0.77$, exceeding those of both FCI evaluation protocols across all three sample sizes. This indicates that PCG-CD recovers a larger proportion of true directed relationships under the considered settings. 

For graph similarity, PCG-CD achieves Sim1 values between $0.91$ and $0.92$ and Sim2 values between $0.84$ and $0.88$. These values are comparable to those of the FCI variants, although their relative performance differs across sample sizes and metrics. Overall, the results indicate that PCG-CD provides useful recovery of directed causal relations, while accurate identification of latent-confounded pairs remains difficult. Since different random graph instances are used for different sample sizes, the observed differences should not be interpreted as effects of sample size alone. Additional comparisons of the decision order (whether to evaluate the no-edge model or apply the $\Delta$ criterion first) and kernel specification under the same fixed threshold are provided in the Appendix \ref{app:exp3_decision_order} and \ref{app:exp3_kernel}, respectively.

\paragraph{Effect of graph density.}

To further investigate the effect of graph density on PCG-CD, we conduct an additional multi-node experiment with explicitly controlled numbers of edges. Each graph contains five observed nodes and one latent node $Z$. The latent node has exactly two observed children, and no direct causal edge is introduced between these two children. We consider three graph-density settings, denoted as e4, e6, and e8, with 4, 6, and 8 total edges, respectively. Since two edges connect the latent node to its observed children, the numbers of edges among the observed nodes are 2, 4, and 6, respectively.

We use nonlinear causal mechanisms based on cubic polynomials and fix the sample size at $n=500$. For each density setting, we evaluate 20 independently generated datasets. All experiments use the same PCG-CD configuration as in the preceding multi-node experiment, including z-score normalization and the same kernel specification. No parameters are adjusted separately for different graph densities.

In addition to Z-hit, we introduce several metrics to evaluate different aspects of graph recovery. \textbf{Confounding Precision} is the proportion of predicted latent-confounded pairs that correspond to the true confounded pair. \textbf{$\operatorname{rank}(\Delta_Z)$} denotes the rank of the true confounded pair when all ten unordered pairs of observed variables are sorted in ascending order of $\Delta$, with smaller ranks indicating stronger pseudo-collinearity. The ranking uses the directional code-length differences from the final graph-update stage before cycle removal, with tied values assigned their average rank. \textbf{$\Delta_Z$ Top-3} is the proportion of datasets in which the true confounded pair is ranked within the three smallest $\Delta$ values.

\textbf{Directed Recall} measures the proportion of true directed edges that are correctly recovered. \textbf{Direct Precision} measures the proportion of estimated directed edges that correspond exactly to true direct causal edges. We also introduce Ancestry-Consistent Precision (\textbf{ACP}), which measures the proportion of estimated directed edges whose directions agree with the true ancestor-descendant relationships, including both direct and indirect ancestry. The difference between ACP and Direct Precision therefore reflects predicted directed edges that are not true direct edges but are consistent with indirect causal ancestry.

Furthermore, $|E_{\mathrm{est}}|$ denotes the number of estimated adjacencies among the observed variables, counting each connected unordered pair once. Macro-F1 is calculated by first pooling the pairwise prediction counts across datasets and then averaging the F1 scores for the no-edge, confounded, and directed relation classes. For Confounding Precision, Directed Recall, Direct Precision, and ACP, the reported values are pooled ratios. In contrast, $\operatorname{rank}(\Delta_Z)$ and $|E_{\mathrm{est}}|$ are reported as the mean and standard deviation across the 20 datasets.

\begin{table}[t]
\caption{Effect of graph density on latent-confounder recovery, directed causal structure recovery, and graph estimation under nonlinear mechanisms with $n=500$.}
\label{tab:exp4_density}
\centering
\begin{tabular}{lccc}
\toprule
Metric & e4 & e6 & e8 \\
\midrule
Z-hit & 0.350 & 0.400 & 0.500 \\
Confounding Precision & 0.318 & 0.154 & 0.154 \\
$\operatorname{rank}(\Delta_Z)$ & $6.400\pm2.664$ & $4.150\pm2.477$ & $4.450\pm2.743$ \\
$\Delta_Z$ Top-3 & 0.200 & 0.500 & 0.450 \\
Directed Recall & 0.675 & 0.512 & 0.358 \\
Direct Precision & 0.458 & 0.612 & 0.632 \\
ACP & 0.458 & 0.627 & 0.647 \\
$|E_{\mathrm{est}}|$ & $4.050\pm1.146$ & $5.950\pm2.282$ & $6.650\pm1.631$ \\
Macro-F1 & 0.584 & 0.492 & 0.420 \\
\botrule
\end{tabular}
\end{table}

Table~\ref{tab:exp4_density} summarizes the results under the three graph-density settings. Z-hit increases from $0.350$ for e4 to $0.500$ for e8, indicating that the true latent-confounded pair is recovered more frequently in the denser settings. However, Confounding Precision decreases from $0.318$ to $0.154$, showing that the increased recovery of true confounded pairs is accompanied by a larger proportion of incorrect confounding predictions. 

The results for $\operatorname{rank}(\Delta_Z)$ and $\Delta_Z$ Top-3 show a different pattern. The average rank improves from $6.400$ for e4 to $4.150$ for e6, but increases to $4.450$ for e8. Similarly, $\Delta_Z$ Top-3 increases from $0.200$ to $0.500$ and then decreases slightly to $0.450$. These results do not indicate a monotonic dependence of $\Delta$-pseudo-collinearity on graph density. 

For directed causal relationships, Directed Recall decreases from $0.675$ for e4 to $0.358$ for e8, indicating that a smaller proportion of true directed edges is recovered as graph density increases. In contrast, Direct Precision increases from $0.458$ to $0.632$. ACP is only slightly higher than Direct Precision, with differences of $0.000$, $0.015$, and $0.015$ for e4, e6, and e8, respectively. This indicates that estimated edges consistent with indirect causal ancestry account for only a small proportion of the predictions in these experiments.

The average number of estimated adjacencies increases from $4.050$ to $6.650$ as graph density increases. The corresponding numbers of true observed adjacencies are 3, 5, and 7 (including the latent confounding pair). Thus, the estimated graphs are, on average, slightly more connected than the true graphs for e4 and e6, while the estimated adjacency count is slightly lower than the true count for e8. This suggests that the complexity penalty in the LNML code-length may help prevent the estimated graphs from becoming excessively dense.

Finally, Macro-F1 decreases from $0.584$ for e4 to $0.420$ for e8, indicating lower overall relation-classification performance in the denser settings under the fixed configuration. Taken together, these results show that graph density affects latent-confounder recovery and directed causal structure recovery in different ways. Although denser graphs yield higher Z-hit and Direct Precision in the present experiments, they also exhibit lower Confounding Precision, Directed Recall, and Macro-F1. 

\subsection{Real-world Data}
We evaluate the proposed method on the Auto MPG dataset\footnote{\url{https://archive.ics.uci.edu/dataset/9/auto+mpg}}, a widely used real-world benchmark for regression and causal analysis. After standard preprocessing, we remove six samples with missing values. We select \emph{cylinders}, \emph{displacement}, \emph{horsepower}, \emph{weight}, and \emph{acceleration} as observable variables, and treat the remaining attributes (\emph{mpg}, \emph{model year}, \emph{origin}, and \emph{car name}) as unobservable.

We apply PCG-CD and FCI to the observable variables, and include ICA-LiNGAM as a reference method applied to the full dataset excluding \emph{car name}. The causal relations and edge strengths inferred by ICA-LiNGAM among the selected variables are extracted for comparison.

\begin{figure}[t]
    \centering
    \begin{subfigure}[t]{0.9\linewidth}
        \centering
        \begin{tikzpicture}[scale=0.3][
    \tikzstyle{every node}=[font=\normalsize]

\definecolor{confblue}{RGB}{70,160,215}

\draw [ fill={rgb,255:red,0; green,0; blue,0} ] (-32.5,-4.25) circle (0.25cm);
\draw  (-32.25,-4.25) circle (0cm);
\draw [ fill={rgb,255:red,0; green,0; blue,0} ] (-32.5,-4.25) circle (0cm);
\node [font=\footnotesize] at (-32,-4.75) {};
\node [font=\footnotesize] at (-31.5,-5.25) {};
\draw [ fill={rgb,255:red,0; green,0; blue,0} ] (-32.5,-4.25) circle (0cm);
\draw [ fill={rgb,255:red,0; green,0; blue,0} ] (-32.5,-4.25) circle (0cm);
\draw [ fill={rgb,255:red,0; green,0; blue,0} ] (-36.25,-7) circle (0.25cm);
\draw [ fill={rgb,255:red,0; green,0; blue,0} ] (-28.75,-7) circle (0.25cm);
\draw [ fill={rgb,255:red,0; green,0; blue,0} ] (-34.5,-10.75) circle (0.25cm);
\draw [ fill={rgb,255:red,0; green,0; blue,0} ] (-30.5,-10.75) circle (0.25cm);

\node [font=\tiny] at (-33,-3.5) {cylinder};
\node [font=\tiny] at (-38.5,-6) {displacement};
\node [font=\tiny] at (-34.5,-11.5) {horsepower};
\node [font=\tiny] at (-30,-11.5) {weight};
\node [font=\tiny] at (-26.25,-6) {acceleration};
\node [font=\tiny] at (-28.5,-7.5) {};

\draw [->, >=Stealth, black] (-36,-6.75) -- (-32.5,-4.25);

\draw [->, >=Stealth, confblue, dashed] (-34.5,-10.5) -- (-32.5,-4.25);

\draw [->, >=Stealth, confblue, dashed] (-36.25,-7.25) -- (-34.5,-10.5);

\draw [->, >=Stealth, confblue, dashed] (-28.75,-7) -- (-36,-7);

\draw [->, >=Stealth, confblue, dashed] (-34.25,-10.5) -- (-28.75,-7);

\draw [->, >=Stealth, confblue, dashed] (-30.75,-10.5) -- (-36,-7.25);

\draw [->, >=Stealth, confblue, dashed] (-34.25,-10.75) -- (-30.75,-10.75);

\draw [->, >=Stealth, confblue, dashed] (-28.75,-7.25) -- (-30.25,-10.5);

\node [font=\footnotesize] at (-32,-4.75) {};
\node [font=\footnotesize] at (-32,-4.75) {};

\draw [ color={rgb,255:red,145; green,145; blue,145} , fill={rgb,255:red,145; green,145; blue,145}] (-20.25,-4.5) circle (0.25cm);
\draw [ color={rgb,255:red,145; green,145; blue,145} , fill={rgb,255:red,145; green,145; blue,145}] (-21.25,-6.75) circle (0.25cm);
\draw [ color={rgb,255:red,145; green,145; blue,145} , fill={rgb,255:red,145; green,145; blue,145}] (-20.25,-9) circle (0.25cm);
\node [font=\footnotesize, color={rgb,255:red,145; green,145; blue,145}] at (-19.75,-10) {};
\draw [ color={rgb,255:red,145; green,145; blue,145} , fill={rgb,255:red,145; green,145; blue,145}] (-21.25,-10.75) circle (0.25cm);

\node [font=\tiny, color={rgb,255:red,145; green,145; blue,145}] at (-18.75,-4.5) {mpg};
\node [font=\tiny, color={rgb,255:red,145; green,145; blue,145}] at (-18,-6.75) {model year};
\node [font=\tiny, color={rgb,255:red,145; green,145; blue,145}] at (-18.25,-9) {origin};
\node [font=\tiny, color={rgb,255:red,145; green,145; blue,145}] at (-18.75,-10.75) {car name};

\node [font=\tiny] at (-35,-5.5) {238.53};
\node [font=\tiny] at (-35.75,-9.25) {11.04};
\node [font=\tiny] at (-34.75,-7.75) {18.21};
\node [font=\tiny] at (-33,-6.5) {10.67};
\node [font=\tiny] at (-31.5,-5.5) {10.92};
\node [font=\tiny] at (-30,-7.75) {86.63};
\node [font=\tiny] at (-29,-9.2) {2.31};
\node [font=\tiny] at (-32.5,-10.3) {11.77};

\end{tikzpicture}
        \caption{Causal graph inferred by PCG-CD. Black solid arrows indicate directed causal relations. Blue dashed arrows indicate potential latent-confounded pairs, with arrowheads pointing in the direction of the shorter code-length. Edge labels indicate $\Delta$.}
        \label{fig:PCG-CD}
    \end{subfigure}\\
    \vspace{0.8em}
    \makebox[\linewidth][c]{%
        \begin{subfigure}[t]{0.48\linewidth}
            \centering
            \begin{tikzpicture}[scale=0.3][
    \tikzstyle{every node}=[font=\tiny]
\draw [ fill={rgb,255:red,0; green,0; blue,0} ] (-32.5,-4.25) circle (0.25cm);
\draw  (-32.25,-4.25) circle (0cm);
\draw [ fill={rgb,255:red,0; green,0; blue,0} ] (-32.5,-4.25) circle (0cm);
\node [font=\footnotesize] at (-32,-4.75) {};
\node [font=\footnotesize] at (-31.5,-5.25) {};
\draw [ fill={rgb,255:red,0; green,0; blue,0} ] (-32.5,-4.25) circle (0cm);
\draw [ fill={rgb,255:red,0; green,0; blue,0} ] (-32.5,-4.25) circle (0cm);
\draw [ fill={rgb,255:red,0; green,0; blue,0} ] (-36.25,-7) circle (0.25cm);
\draw [ fill={rgb,255:red,0; green,0; blue,0} ] (-28.75,-7) circle (0.25cm);
\draw [ fill={rgb,255:red,0; green,0; blue,0} ] (-34.5,-10.75) circle (0.25cm);
\draw [ fill={rgb,255:red,0; green,0; blue,0} ] (-30.5,-10.75) circle (0.25cm);
\node [font=\tiny] at (-33,-3.5) {cylinder};
\node [font=\tiny] at (-37,-6) {displacement};
\node [font=\tiny] at (-34.5,-11.5) {horsepower};
\node [font=\tiny] at (-30,-11.5) {weight};
\node [font=\tiny] at (-29,-6) {acceleration};
\node [font=\normalsize] at (-28.5,-7.5) {};
\node [font=\footnotesize] at (-32,-4.75) {};
\node [font=\footnotesize] at (-32,-4.75) {};
\node [font=\footnotesize, color={rgb,255:red,145; green,145; blue,145}] at (-19.75,-10) {};
\draw [->, >=Stealth] (-36.25,-7) -- (-32.75,-4.5);
\draw [->, >=Stealth, dashed] (-29,-7) -- (-36,-7);
\draw [->, >=Stealth, dashed] (-28.75,-7) -- (-34.25,-10.5);
\draw [->, >=Stealth, dashed] (-30.5,-10.75) -- (-36,-7.25);
\draw [->, >=Stealth, dashed] (-30.5,-10.75) -- (-34.25,-10.75);
\draw [<->, >=Stealth, dashed] (-36.25,-7.25) -- (-34.75,-10.5);
\end{tikzpicture}
            \caption{Partial ancestral graph inferred by FCI. Dashed edges indicate partially oriented edges.}
            \label{Fig:FCI}
        \end{subfigure}
        \begin{subfigure}[t]{0.48\linewidth}
            \centering
            \begin{tikzpicture}[scale=0.3][
    \tikzstyle{every node}=[font=\tiny]
\draw [ fill={rgb,255:red,0; green,0; blue,0} ] (-32.5,-4.25) circle (0.25cm);
\draw  (-32.25,-4.25) circle (0cm);
\draw [ fill={rgb,255:red,0; green,0; blue,0} ] (-32.5,-4.25) circle (0cm);
\node [font=\footnotesize] at (-32,-4.75) {};
\node [font=\footnotesize] at (-31.5,-5.25) {};
\draw [ fill={rgb,255:red,0; green,0; blue,0} ] (-32.5,-4.25) circle (0cm);
\draw [ fill={rgb,255:red,0; green,0; blue,0} ] (-32.5,-4.25) circle (0cm);
\draw [ fill={rgb,255:red,0; green,0; blue,0} ] (-36.25,-7) circle (0.25cm);
\draw [ fill={rgb,255:red,0; green,0; blue,0} ] (-28.75,-7) circle (0.25cm);
\draw [ fill={rgb,255:red,0; green,0; blue,0} ] (-34.5,-10.75) circle (0.25cm);
\draw [ fill={rgb,255:red,0; green,0; blue,0} ] (-30.5,-10.75) circle (0.25cm);
\node [font=\tiny] at (-33,-3.5) {cylinder};
\node [font=\tiny] at (-37.5,-6) {displacement};
\node [font=\tiny] at (-34.5,-11.5) {horsepower};
\node [font=\tiny] at (-30,-11.5) {weight};
\node [font=\tiny] at (-28,-6) {acceleration};
\node [font=\normalsize] at (-28.5,-7.5) {};
\draw [->, >=Stealth] (-32.5,-4.25) -- (-36,-6.75);
\draw [->, >=Stealth] (-32.5,-4.25) -- (-34.5,-10.5);
\draw [->, >=Stealth] (-36,-7.25) -- (-30.75,-10.5);
\draw [->, >=Stealth] (-34.25,-10.75) -- (-30.75,-10.75);
\draw [->, >=Stealth] (-28.75,-7.25) -- (-30.25,-10.5);
\draw [->, >=Stealth] (-28.75,-7) -- (-34.25,-10.5);
\node [font=\footnotesize] at (-32,-4.75) {};
\node [font=\footnotesize] at (-32,-4.75) {};
\node [font=\footnotesize, color={rgb,255:red,145; green,145; blue,145}] at (-19.75,-10) {};
\draw [->, >=Stealth] (-34.75,-10.5) -- (-36.25,-7.25);
\node [font=\tiny] at (-35,-5.5) {36.4};
\node [font=\tiny] at (-35.75,-9.25) {0.93};
\node [font=\tiny] at (-34.75,-7.75) {4.63};
\node [font=\tiny] at (-33,-6.5) {10.1};
\node [font=\tiny] at (-30.5,-7.75) {-4.7};
\node [font=\tiny] at (-29,-8.5) {60.5};
\node [font=\tiny] at (-32.5,-10.2) {7.16};
\draw [->, >=Stealth] (-32.5,-4.25) -- (-29,-6.75);
\node [font=\tiny] at (-30.25,-5.25) {-0.8};
\end{tikzpicture}
            \caption{Causal graph inferred by ICA-LiNGAM. The values indicate estimated edge strengths.}
            \label{fig:LiNGAM}
        \end{subfigure}%
    }
    \caption{Causal discovery results on the Auto MPG dataset.}
    \label{fig:auto_mpg}
\end{figure}
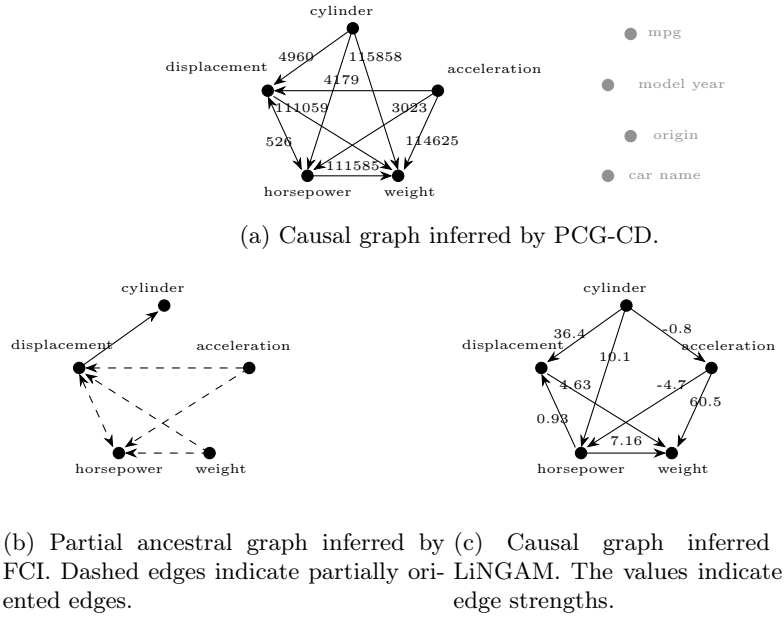
Figure~\ref{fig:auto_mpg} compares the causal graphs inferred by different methods. For PCG-CD, we use the same data standardization, kernel specification, and threshold as in the synthetic experiments to maintain a consistent experimental setting, although this configuration is not necessarily optimal for the Auto MPG dataset. The estimated graph contains one directed edge and seven potential latent-confounded pairs. Although PCD-CD algorithm adopt the no-edge-first decision rule to avoid unnecessary confounding predictions, it does not necessarily result in a small number of potential confounded pairs.

Since the ground-truth causal graph is unavailable, we compare the inferred relations with those obtained by FCI and ICA-LiNGAM. Both PCG-CD and FCI leave \emph{cylinder} and \emph{acceleration} disconnected and identify a directed edge from \emph{displacement} to \emph{cylinder}, for which PCG-CD obtains a relatively large $\Delta$ of $238.53$. Although FCI leaves \emph{acceleration} and \emph{weight} disconnected, PCG-CD assigns a potential confounded relation with a very small $\Delta$ of $2.31$, indicating little preference between the two causal directions. Furthermore, for the \emph{displacement}--\emph{acceleration} and \emph{displacement}--\emph{weight} pairs, the directions with shorter code-lengths are consistent with the arrowhead orientations of the corresponding partially oriented edges in FCI. Similar directional preferences are observed for \emph{horsepower}--\emph{weight} and \emph{acceleration}--\emph{weight} when compared with ICA-LiNGAM. Although different causal discovery methods may produce different causal graphs due to their underlying assumptions, the directional preferences reflected by the code-length differences in PCG-CD show consistency with certain relations inferred by FCI and ICA-LiNGAM, providing complementary information about the underlying causal structure.

\section{Conclusion}
In this paper, we proposed an MDL-based framework for causal discovery that accommodates both nonlinear causal mechanisms and latent confounders. By leveraging LNML code-lengths, the proposed method evaluates causal relationships between variable pairs among four possible cases: $A \rightarrow B$, $A \leftarrow B$, $A \leftrightarrow B$, and $A \nleftrightarrow B$, and reconstructs the causal graph by minimizing the overall description length.

A key component of our approach is the notion of \(\Delta\)-pseudo-collinearity, which provides a criterion for identifying potential latent-confounding relations. We theoretically characterize this criterion under general nonlinear mechanisms, showing that the directional code-length gap grows at most linearly with the sample size under regularity conditions and decreases as the latent causal mechanisms become more similar. Building on these results, we developed PCG-CD, a greedy algorithm that achieves lower edge-search complexity than conventional score-based methods.

Experimental results on synthetic data demonstrate the usefulness of $\Delta$-pseudo-collinearity for flagging potential latent-confounding candidates under nonlinear mechanisms and show that its behavior depends on threshold choice and graph structure. Building on this criterion, PCG-CD recovers directed causal relations across different synthetic settings, while the graph-density experiment shows that latent-confounder recovery, confounding precision, and directed-edge recall respond differently to increasing graph density. On the real-world Auto MPG dataset, certain directional preferences reflected by the code-length differences in PCG-CD are consistent with relations inferred by FCI and ICA-LiNGAM, providing complementary information about the underlying causal structure.

\section*{Declarations}

\begin{appendices}

\section{Proof of Theorem~\ref{the:general_nonlinear}}\label{apd:general}
\begin{proof}
Let
\[
A=g_A(Z)+\epsilon_A,\qquad
B=g_B(Z)+\epsilon_B,
\]
where
\[
Z\sim\mathcal{N}(\mu,\sigma_Z^2),\qquad
\epsilon_A\sim\mathcal{N}(0,\sigma_A^2),\qquad
\epsilon_B\sim\mathcal{N}(0,\sigma_B^2),
\]
and $Z$, $\epsilon_A$, and $\epsilon_B$ are mutually independent.

According to the definition of the LNML code-length, the difference between
the two directional code-lengths can be written as
\[
\left|\Delta\right|
=
\left|
\Delta_{\mathrm{marginal}}
+\Delta_{\mathrm{fit}}
+\Delta_{\mathrm{comp}}
+\Delta_{\mathrm{norm}}
\right|,
\]
where
\[
\Delta_{\mathrm{marginal}}
=
-\log P(A)+\log P(B),
\]
\[
\Delta_{\mathrm{fit}}
=
-\log P_{\mathrm{GP}}(B\mid A)
+\log P_{\mathrm{GP}}(A\mid B),
\]
\[
\Delta_{\mathrm{comp}}
=
\|f_{B\mid A}\|_{\mathcal H_k}^{2}
-
\|f_{A\mid B}\|_{\mathcal H_k}^{2},
\]
and
\[
\Delta_{\mathrm{norm}}
=
\frac{1}{2}\log\det(\sigma^{-2}K_A+I)
-
\frac{1}{2}\log\det(\sigma^{-2}K_B+I).
\]

We first consider the relation between the two observed variables.
On the effective support of $Z$, $g_A$ and $g_B$ are locally invertible.
Let
\[
h_A=g_A^{-1},\qquad h_B=g_B^{-1}.
\]
Ignoring the observation noise temporarily gives
\[
B=g_B(h_A(A))=:F(A),\qquad
A=g_A(h_B(B))=:G(B),
\]
where $G=F^{-1}$ locally.

Taking the noise terms into account, from
\[
A=g_A(Z)+\epsilon_A
\]
we have
\[
Z=h_A(A-\epsilon_A).
\]
Using a first-order Taylor expansion,
\[
h_A(A-\epsilon_A)
\approx
h_A(A)-h_A'(A)\epsilon_A.
\]
Hence
\[
B
\approx
F(A)
-
\frac{g_B'(h_A(A))}
     {g_A'(h_A(A))}
\epsilon_A
+\epsilon_B.
\]
Therefore,
\[
B=F(A)+\eta_{B\mid A},
\]
with
\[
\eta_{B\mid A}
\approx
-
\frac{g_B'(h_A(A))}
     {g_A'(h_A(A))}
\epsilon_A+\epsilon_B.
\]
Similarly,
\[
A=G(B)+\eta_{A\mid B},
\]
where
\[
\eta_{A\mid B}
\approx
-
\frac{g_A'(h_B(B))}
     {g_B'(h_B(B))}
\epsilon_B+\epsilon_A.
\]

Thus, under the local invertibility assumption, both directions can be
represented by smooth regression functions with bounded effective noise.
Since the first derivatives of $g_A$ and $g_B$ are bounded away from zero and
infinity, the corresponding conditional variances
\[
v_{B\mid A}(a)
\approx
\left(
\frac{g_B'(h_A(a))}
     {g_A'(h_A(a))}
\right)^2
\sigma_A^2+\sigma_B^2
\]
and
\[
v_{A\mid B}(b)
\approx
\left(
\frac{g_A'(h_B(b))}
     {g_B'(h_B(b))}
\right)^2
\sigma_B^2+\sigma_A^2
\]
remain bounded.

Let
\[
m_A=\mathbb E[A],\qquad
m_B=\mathbb E[B],
\]
and
\[
v_A=\operatorname{Var}(A),\qquad
v_B=\operatorname{Var}(B).
\]
Under the Gaussian moment-matching approximation used for the marginal
terms,
\[
A\approx\mathcal{N}(m_A,v_A),\qquad
B\approx\mathcal{N}(m_B,v_B).
\]
Then
\[
\Delta_{\mathrm{marginal}}
=
\frac{n}{2}\log\frac{v_A}{v_B}
+
\sum_{i=1}^{n}
\left(
\frac{(a_i-m_A)^2}{2v_A}
-
\frac{(b_i-m_B)^2}{2v_B}
\right).
\]
Likewise, using the local regression representation above,
\[
\Delta_{\mathrm{fit}}
\approx
\frac{1}{2}
\sum_{i=1}^{n}
\left[
\log
\frac{v_{B\mid A}(a_i)}
     {v_{A\mid B}(b_i)}
+
\frac{(b_i-F(a_i))^2}
     {v_{B\mid A}(a_i)}
-
\frac{(a_i-G(b_i))^2}
     {v_{A\mid B}(b_i)}
\right].
\]
Under the boundedness assumptions, each summand is bounded independently
of $n$. Therefore,
\[
\left|
\Delta_{\mathrm{marginal}}
+\Delta_{\mathrm{fit}}
\right|
\leq nC_{\mathrm{mf}},
\]
where $C_{\mathrm{mf}}$ is independent of $n$.
The constant $C_{\mathrm{mf}}$ becomes small as $g_A$ and $g_B$ become more
similar, since the induced marginal distributions and the two local conditional
regressions become closer. It vanishes when the corresponding marginal and
conditional terms coincide.

We next consider the RKHS complexity term. Since $f_{B\mid A}$ minimizes
the corresponding GP code-length objective, for any $f\in\mathcal H_k$,
\[
-\log P_{\mathrm{GP}}(B\mid A;f_{B\mid A})
+\|f_{B\mid A}\|_{\mathcal H_k}^{2}
\leq
-\log P_{\mathrm{GP}}(B\mid A;f)
+\|f\|_{\mathcal H_k}^{2}.
\]
Taking $F=g_B\circ g_A^{-1}$ as the reference function gives
\[
\|f_{B\mid A}\|_{\mathcal H_k}^{2}
\leq
-\log P_{\mathrm{GP}}(B\mid A;F)
+\|F\|_{\mathcal H_k}^{2}.
\]
Similarly,
\[
\|f_{A\mid B}\|_{\mathcal H_k}^{2}
\leq
-\log P_{\mathrm{GP}}(A\mid B;G)
+\|G\|_{\mathcal H_k}^{2}.
\]
Since $F$ and $G$ are smooth on the effective support and their RKHS norms
are bounded independently of $n$, while the corresponding residual
log-likelihoods are of order $n$, there exists a constant
$C_{\mathrm{comp}}$ independent of $n$ such that
\[
|\Delta_{\mathrm{comp}}|
\leq nC_{\mathrm{comp}}.
\]
The constant $C_{\mathrm{comp}}$ becomes small when the two local regression
functions $F=g_B\circ g_A^{-1}$ and $G=g_A\circ g_B^{-1}$ have comparable
complexity. In particular, it approaches zero when the two directional RKHS
complexities are similar.

Finally, consider the normalization term. Writing it in terms of the
eigenvalues of the kernel matrices,
\[
\Delta_{\mathrm{norm}}
=
\frac{1}{2}
\sum_{r=1}^{n}
\left[
\log\left(1+\sigma^{-2}\lambda_r(K_A)\right)
-
\log\left(1+\sigma^{-2}\lambda_r(K_B)\right)
\right].
\]
By the comparability assumption in Theorem \ref{the:general_nonlinear}, there exist constants
$c_1,c_2>0$, independent of $n$, such that
\[
c_1\lambda_r(K_A)
\leq
\lambda_r(K_B)
\leq
c_2\lambda_r(K_A),
\qquad r=1,\ldots,n.
\]
For $x=\sigma^{-2}\lambda_r(K_A)\geq0$, we then have
\[
\sigma^{-2}\lambda_r(K_B)\in[c_1x,c_2x].
\]
Hence
\[
\left|
\log(1+x)
-
\log\left(1+\sigma^{-2}\lambda_r(K_B)\right)
\right|
\leq C_0,
\]
where
\[
C_0=\max\{\log c_2,-\log c_1\}.
\]
Therefore,
\[
|\Delta_{\mathrm{norm}}|
\leq
\frac{n}{2}C_0
=
nC_{\mathrm{norm}}.
\]
The constant $C_{\mathrm{norm}}$ becomes small as the spectra of $K_A$ and
$K_B$ become closer. In the limiting case where their eigenvalues coincide,
$C_{\mathrm{norm}}=0$.

Combining the three bounds gives
\[
\Delta
\leq
n\left(
C_{\mathrm{mf}}
+
C_{\mathrm{comp}}
+
C_{\mathrm{norm}}
\right)
=
nC(g_A,g_B,\sigma_Z,\sigma_A,\sigma_B).
\]
As $g_A$ and $g_B$ become more similar, the corresponding marginal,
regression-complexity, and kernel-spectrum differences all become smaller, so
the three constants above decrease accordingly.
\end{proof}

\section{Proof of the Quadratic Specialization}\label{apd:quadratic}
\begin{proof}
Let
\[
A=g_A(Z)+\epsilon_A,\qquad
B=g_B(Z)+\epsilon_B,
\]
where
\[
g_A(z)=\alpha_1+\alpha_2z+\alpha_3z^2,\qquad
g_B(z)=\beta_1+\beta_2z+\beta_3z^2,
\]
and
\[
Z\sim\mathcal{N}(\mu,\sigma_Z^2),\qquad
\epsilon_A\sim\mathcal{N}(0,\sigma_A^2),\qquad
\epsilon_B\sim\mathcal{N}(0,\sigma_B^2),
\]
with $Z$, $\epsilon_A$, and $\epsilon_B$ mutually independent.

To express the contribution of the latent variable explicitly, define
\[
U:=Z-\mu,
\qquad
Q:=U^2-\sigma_Z^2.
\]
Then $\mathbb{E}[U]=\mathbb{E}[Q]=0$, and
\[
Z=\mu+U,\qquad
Z^2=\mu^2+2\mu U+U^2.
\]
Substituting these expressions into $A$ gives
\[
A
=
\alpha_1+\alpha_2(\mu+U)
+\alpha_3(\mu^2+2\mu U+U^2)
+\epsilon_A.
\]
Hence,
\[
A=m_A+c_AU+\alpha_3Q+\epsilon_A,
\]
where
\[
m_A
:=
\alpha_1+\alpha_2\mu
+\alpha_3(\mu^2+\sigma_Z^2),
\qquad
c_A
:=
\alpha_2+2\alpha_3\mu.
\]
Similarly,
\[
B=m_B+c_BU+\beta_3Q+\epsilon_B,
\]
where
\[
m_B
:=
\beta_1+\beta_2\mu
+\beta_3(\mu^2+\sigma_Z^2),
\qquad
c_B
:=
\beta_2+2\beta_3\mu.
\]

Since $U\sim\mathcal{N}(0,\sigma_Z^2)$,
\[
\operatorname{Var}(U)=\sigma_Z^2,
\qquad
\operatorname{Var}(Q)=2\sigma_Z^4,
\qquad
\operatorname{Cov}(U,Q)=0.
\]
Therefore,
\[
\mathbb{E}[A]=m_A,\qquad
\mathbb{E}[B]=m_B,
\]
and
\[
\operatorname{Var}(A)
=
c_A^2\sigma_Z^2
+
2\alpha_3^2\sigma_Z^4
+
\sigma_A^2,
\]
\[
\operatorname{Var}(B)
=
c_B^2\sigma_Z^2
+
2\beta_3^2\sigma_Z^4
+
\sigma_B^2.
\]
Moreover,
\[
\operatorname{Cov}(A,B)
=
c_Ac_B\sigma_Z^2
+
2\alpha_3\beta_3\sigma_Z^4.
\]
Thus, unlike the linear case, the shared latent variation contains both
first-order and second-order components.

We next consider the directional regressions between $A$ and $B$. Under the
local invertibility assumption in Theorem~1, define
\[
h_A:=g_A^{-1},\qquad
h_B:=g_B^{-1}
\]
on the relevant local branch. The corresponding noiseless mappings are
\[
F:=g_B\circ h_A,
\qquad
G:=g_A\circ h_B.
\]
Hence, in the absence of observation noise,
\[
B=F(A),\qquad
A=G(B),
\]
with $G=F^{-1}$ locally.

Taking the observation noise into account, from
\[
A=g_A(Z)+\epsilon_A
\]
we have
\[
Z=h_A(A-\epsilon_A).
\]
Using a first-order Taylor expansion around $A$,
\[
h_A(A-\epsilon_A)
\approx
h_A(A)-h_A'(A)\epsilon_A.
\]
Since
\[
h_A'(a)
=
\frac{1}{g_A'(h_A(a))},
\]
substituting into $B=g_B(Z)+\epsilon_B$ gives
\[
B
\approx
F(A)
-
\frac{g_B'(h_A(A))}
     {g_A'(h_A(A))}
\epsilon_A
+
\epsilon_B.
\]
Thus,
\[
B=F(A)+\eta_{B\mid A},
\]
where
\[
\eta_{B\mid A}
\approx
-
\frac{g_B'(h_A(A))}
     {g_A'(h_A(A))}
\epsilon_A
+
\epsilon_B.
\]
Similarly,
\[
A=G(B)+\eta_{A\mid B},
\]
where
\[
\eta_{A\mid B}
\approx
-
\frac{g_A'(h_B(B))}
     {g_B'(h_B(B))}
\epsilon_B
+
\epsilon_A.
\]

The corresponding local conditional variances are therefore
\[
v_{B\mid A}(a)
\approx
\left(
\frac{g_B'(h_A(a))}
     {g_A'(h_A(a))}
\right)^2
\sigma_A^2
+
\sigma_B^2,
\]
and
\[
v_{A\mid B}(b)
\approx
\left(
\frac{g_A'(h_B(b))}
     {g_B'(h_B(b))}
\right)^2
\sigma_B^2
+
\sigma_A^2.
\]
These are the quadratic counterparts of the conditional variances appearing
in the general nonlinear case.

Recall that
\[
\Delta
=
\left|
\Delta_{\mathrm{marginal}}
+
\Delta_{\mathrm{fit}}
+
\Delta_{\mathrm{comp}}
+
\Delta_{\mathrm{norm}}
\right|.
\]
Using the marginal variances derived above and the same Gaussian
moment-matching approximation as in Theorem~1,
\[
\Delta_{\mathrm{marginal}}
=
\frac{n}{2}
\log
\frac{\operatorname{Var}(A)}
     {\operatorname{Var}(B)}
+
\sum_{i=1}^{n}
\left[
\frac{(a_i-m_A)^2}
     {2\operatorname{Var}(A)}
-
\frac{(b_i-m_B)^2}
     {2\operatorname{Var}(B)}
\right].
\]
The directional fit term is approximated by
\[
\Delta_{\mathrm{fit}}
\approx
\frac{1}{2}
\sum_{i=1}^{n}
\left[
\log
\frac{v_{B\mid A}(a_i)}
     {v_{A\mid B}(b_i)}
+
\frac{(b_i-F(a_i))^2}
     {v_{B\mid A}(a_i)}
-
\frac{(a_i-G(b_i))^2}
     {v_{A\mid B}(b_i)}
\right].
\]
Under the boundedness assumptions in Theorem~\ref{the:general_nonlinear}, these terms are bounded
independently of $n$, and therefore
\[
\left|
\Delta_{\mathrm{marginal}}
+
\Delta_{\mathrm{fit}}
\right|
\leq
nC_{\mathrm{mf}},
\]
for some constant $C_{\mathrm{mf}}$ independent of $n$.
For the quadratic mechanisms considered here, similarity between $g_A$ and $g_B$ corresponds to closeness of their polynomial coefficients, in particular the linear and quadratic coefficients $\alpha_2,\beta_2$ and $\alpha_3,\beta_3$. As these coefficients become closer, the induced marginal moments and local conditional regressions become more similar, and $C_{\mathrm{mf}}$ decreases accordingly.

The dependence of $C_{\mathrm{mf}}$ on $\sigma_Z^2$ is generally non-monotonic. Increasing $\sigma_Z^2$ strengthens the shared latent variation, but also amplifies the quadratic components through the $\sigma_Z^4$ terms in the marginal variances and covariance, while changing the local derivative ratios in the conditional variances. These effects need not reduce the two directional code-lengths at the same rate.

The bounds for the RKHS complexity and normalization terms follow directly
from the argument of Theorem~\ref{the:general_nonlinear}:
\[
|\Delta_{\mathrm{comp}}|
\leq nC_{\mathrm{comp}},
\qquad
|\Delta_{\mathrm{norm}}|
\leq nC_{\mathrm{norm}},
\]
where both constants are independent of $n$. As in the general nonlinear case, $C_{\mathrm{comp}}$ becomes small when the two directional regression functions have similar RKHS complexity, while $C_{\mathrm{norm}}$ decreases as the spectra of the corresponding kernel matrices become closer.
\end{proof}

\section{Proof of Corollary~\ref{cor:linear}}\label{apd:linear}

\begin{proof}
Denote the values of nodes $A$ and $B$ as $a$ and $b$, respectively. Assume
\begin{align}
    & Z \sim \mathcal{N}(\mu,\sigma_Z^2), \nonumber\\
    & a = \alpha_1 +\alpha_2 z+ \epsilon_A, \quad \epsilon_A \sim \mathcal{N}(0,\sigma_A^2), \nonumber\\
    & b = \beta_1 +\beta_2 z+ \epsilon_B, \quad \epsilon_B \sim \mathcal{N}(0,\sigma_B^2). \nonumber
\end{align}
The latent relations in $a$ and $b$ are unobserved.

Consider first the directional code-length in \eqref{eq:dpcl} corresponding to $A\to B$. Since node $A$ has no parents, its marginal term becomes
\begin{equation}
    L(A_{\text{res}}|\text{pa}_{A_{\text{res}}}=\emptyset) = -\log P(A), \nonumber
\end{equation}
where $-\log P(A)$ is the negative log-likelihood of $A$ under the MLE
\begin{align}
    \hat{\mu}_A &= \alpha_1+\alpha_2\mu, \nonumber\\
    \hat{\sigma}_A^2 &= \alpha_2^2\sigma_Z^2+\sigma_A^2. \nonumber
\end{align}

The second term $L(B_{\text{res}}|\text{pa}_{B_{\text{res}}}=A_{\text{res}})$ depends on the relation between $B$ and $A$. The latent linear relation can be rewritten as
\begin{equation}
    b = \Big(\beta_1-\tfrac{\alpha_1}{\alpha_2}\beta_2\Big) + \tfrac{\beta_2}{\alpha_2}a + \Big(\epsilon_B- \tfrac{\beta_2}{\alpha_2}\epsilon_A\Big). \nonumber
\end{equation}
For notational simplicity, let
\begin{align}
    & b = \gamma_1+\gamma_2 a+ \epsilon_B', \quad\epsilon_B' \sim \mathcal{N}(0,\sigma_B'^2), \nonumber\\
    & a \sim \mathcal{N}(\mu_A,\sigma_A^2), \quad b \sim \mathcal{N}(\gamma_1+\gamma_2 \mu_A,\gamma_2^2\sigma_A^2+\sigma_B'^2). \nonumber
\end{align}

Since the true mechanism between $A$ and $B$ is unknown, we assume $B=f(A)+\epsilon_1$ follows a Gaussian Process prior:
\begin{align}
    f \sim \mathcal{GP}(m(a),\kappa(a,a')). \nonumber
\end{align}
Then $\bm{b}|\bm{a}$ follows a multivariate Gaussian distribution with covariance $\bm{\Sigma}_1 = \bm{K}_1+\sigma^2\bm{I}$:
\begin{align}
    \bm{b}|\bm{a} \sim \mathcal{N}(\bm{m}_1,\bm{\Sigma}_1). \nonumber
\end{align}
The corresponding likelihood is
\begin{align}
    -\log P_{\text{GP}}(B|A) = &\tfrac{1}{2}(\bm{b}-\bm{m}_1)^\top \bm{\Sigma}_1^{-1} (\bm{b}-\bm{m}_1) \nonumber \\
    &+ \tfrac{1}{2}\log \det \bm{\Sigma}_1 +\tfrac{n}{2} \log 2\pi, \nonumber
\end{align}
where $n$ is the sample size.

In theory, under a perfect GP regression and assuming the kernel has infinite capacity, the mechanism $f(\cdot)$ can recover the true relationship. In this case,
\begin{align}
    -\log P_{\text{GP}}(B|A) 
    &= \tfrac{n}{2} \log\!\big(2\pi (\gamma_2^2 \sigma_A^2+\sigma_B'^2)\big) \nonumber\\
    &\quad +  \sum_{i=1}^n \frac{(b_i-\gamma_1-\gamma_2 \mu_A)^2}{2(\gamma_2^2 \sigma_A^2+\sigma_B'^2)}. \nonumber
\end{align}
Moreover, as $\bm{K}_1 \to 0$, we have
\begin{equation}
    L(B_{\text{res}}|\text{pa}_{B_{\text{res}}}=A_{\text{res}}) \to -\log P_{\text{GP}}(B|A). \nonumber
\end{equation}

Thus, the code-length for $A\to B$ tends to
\begin{align}
    L(A'_{\text{res}}|\text{pa}_{A'_{\text{res}}}=\emptyset)+&L(B'_{\text{res}}|\text{pa}_{B'_{\text{res}}}=A'_{\text{res}}) \nonumber\\
     &\to -\log P(A)-\log P_{\text{GP}}(B|A). \nonumber
\end{align}
Similarly, the code-length for $B\to A$ tends to
\begin{align}
    L(A'_{\text{res}}|\text{pa}_{A'_{\text{res}}}=B'_{\text{res}})+&L(B'_{\text{res}}|\text{pa}_{B'_{\text{res}}}=\emptyset) \nonumber\\
    &\to -\log P(B)-\log P_{\text{GP}}(A|B). \nonumber
\end{align}

Therefore,
\begin{align}
    \Delta \approx -\log P(A) -\log P_{\text{GP}}(B|A) + \log P(B)+\log P_{\text{GP}}(A|B). \nonumber
\end{align}

\noindent More explicitly, the first part is
\begin{align}
    - \log P(A)+\log P(B) 
    &= \tfrac{n}{2}\log \Big ( \tfrac{\alpha_2^2\sigma_Z^2+\sigma_A^2}{\beta_2^2\sigma_Z^2+\sigma_B^2} \Big ) \nonumber\\
    &\quad + \sum_{i=1}^n \Bigg( \frac{(a_i-\alpha_1-\alpha_2 \mu)^2}{2(\alpha_2^2 \sigma_Z^2+\sigma_A^2)}  - \frac{(b_i-\beta_1-\beta_2\mu)^2}{2(\beta_2^2 \sigma_Z^2+\sigma_B^2)} \Bigg), \nonumber
\end{align}
and the second part is
\begin{align}
    &-\log P_{\text{GP}}(B|A) + \log P_{\text{GP}}(A|B) \nonumber\\
    =& \tfrac{n}{2} \log\!\Bigg( \frac{\beta_2^2\sigma_Z^2 + 2\frac{\beta_2^2}{\alpha_2^2}\sigma_A^2 + \sigma_B^2}{\alpha_2^2\sigma_Z^2+2\frac{\alpha_2^2}{\beta_2^2}\sigma_B^2 + \sigma_A^2} \Bigg) +  \sum_{i=1}^n \Bigg\{ \frac{(b_i-(\beta_1-\tfrac{\alpha_1}{\alpha_2}\beta_2)-\tfrac{\beta_2}{\alpha_2} (\alpha_1+\alpha_2 \mu))^2}{2(\beta_2^2\sigma_Z^2 + 2\frac{\beta_2^2}{\alpha_2^2}\sigma_A^2 + \sigma_B^2)} \nonumber\\
    &\qquad \qquad \qquad \qquad \qquad - \frac{(a_i-(\alpha_1-\tfrac{\beta_1}{\beta_2}\alpha_2)-\tfrac{\alpha_2}{\beta_2} (\beta_1+\beta_2 \mu))^2}{2(\alpha_2^2\sigma_Z^2 + 2\frac{\alpha_2^2}{\beta_2^2}\sigma_B^2 + \sigma_A^2)} \Bigg\}. \nonumber
\end{align}
 
The gap $\Delta$ decreases when $\sigma_Z$ is relatively large, and when $\alpha_2$ is close to $\beta_2$. Moreover, $\Delta$ increases linearly with the data size $n$. Equivalently, within these idealized GP-regression assumptions, it is upper-bounded by $nC(|\alpha_2-\beta_2|,\sigma_Z,\sigma_A,\sigma_B)$ for a constant $C$ independent of $n$.
\end{proof}

\section{Pairwise Robustness and Threshold Sensitivity}
\label{app:pairwise_robustness}

This section examines the sensitivity of the pairwise experiments to data standardization, kernel specification, and the threshold parameter $\epsilon$. The analyses use the same datasets and evaluation metrics as the corresponding experiments in the main text.

\subsection{Effect of Data Standardization and Kernel Specification}
\label{app:pairwise_standardization}

We compare two kernel specifications. The first uses $\mathrm{WhiteKernel}(1.0)$ for variables without parents and $\mathrm{RBF}(1.0)+\mathrm{WhiteKernel}(1.0)$ for variables with parents, which is the same as the main experiments in Section \ref{sec:exp_delta}. The second additionally includes a constant kernel, using $\mathrm{ConstantKernel}(1.0)+\mathrm{WhiteKernel}(1.0)$ for variables without parents and $\mathrm{ConstantKernel}(1.0)\times\mathrm{RBF}(1.0)+\mathrm{WhiteKernel}(1.0)$ for variables with parents. These two specifications are denoted as White and Amplitude, respectively.

\paragraph{Experiment without observable parents.}

Table~\ref{tab:app_exp1_robustness} compares the average AUC scores under the two kernel specifications, with and without data standardization. Each value is the average of the AUC scores across the nine combinations of sample size and noise level, with 20 datasets per setting.

\begin{table}[t]
\caption{Average AUC scores in the pairwise experiment without observable parents under different data preprocessing and kernel specifications.}
\label{tab:app_exp1_robustness}
\centering
\begin{tabular}{llcc}
\toprule
Mechanism & Kernel & Raw data & Standardized data \\
\midrule
\multirow{2}{*}{Linear}
& White & 0.4477 & 0.4607 \\
& Amplitude & 0.7244 & 0.4844 \\
\midrule
\multirow{2}{*}{Nonlinear}
& White & 0.9899 & 0.9849 \\
& Amplitude & 0.7820 & 0.9874 \\
\botrule
\end{tabular}
\end{table}

The results show that data standardization reduces the sensitivity of AUC to the kernel specification. The nonlinear cases achieve average AUC values of $0.9849$ and $0.9874$ after standardization, indicating under both kernel specifications, the discrimination is consistently strong. In contrast, average AUC values of the standardized linear cases are below $0.5$ for both kernels. These results indicate that standardization improves consistency across kernel specifications but does not necessarily improve discrimination in every setting.

\paragraph{Experiment with observable parents.}

Table~\ref{tab:app_exp2_robustness} presents the corresponding results for the pairwise experiment with observable parents. The reported AUC values are averages over the nine sample-size and noise-level combinations. The threshold is fixed at $\epsilon=0.5$.

\begin{table}[t]
\caption{Average AUC and classification accuracy in the pairwise experiment with observable parents under different data preprocessing and kernel specifications.}
\label{tab:app_exp2_robustness}
\centering
\begin{tabular}{lllcccc}
\toprule
Mechanism & Kernel & Data & AUC & Acc1 & Acc2 & Acc3 \\
\midrule
\multirow{4}{*}{Linear}
& White & Raw & 0.9424 & 0.8833 & 0.8333 & 0.8611 \\
& White & Standardized & 0.9653 & 0.8556 & 0.8889 & 0.8944 \\
& Amplitude & Raw & 0.9379 & 0.8889 & 0.8389 & 0.8444 \\
& Amplitude & Standardized & 0.9676 & 0.8556 & 0.8944 & 0.9000 \\
\midrule
\multirow{4}{*}{Nonlinear}
& White & Raw & 0.8792 & 0.4833 & 0.6833 & 0.7944 \\
& White & Standardized & 1.0000 & 0.5333 & 0.9222 & 0.9500 \\
& Amplitude & Raw & 0.6897 & 0.4833 & 0.5611 & 0.6556 \\
& Amplitude & Standardized & 1.0000 & 0.7278 & 0.9333 & 0.9500 \\
\botrule
\end{tabular}
\end{table}

Data standardization improves AUC, Acc2, and Acc3 under both kernel specifications for linear and nonlinear mechanisms. In the nonlinear cases, AUC reaches $1.0000$ for both kernels after standardization. The differences in AUC and Acc3 between the two kernels are also reduced after standardization. However, Acc1 shows a different pattern. In particular, the standardized nonlinear cases yield Acc1 values of $0.5333$ and $0.7278$ under the White and Amplitude kernels, respectively, despite their identical AUC and Acc3 values. This difference indicates that direct four-model code-length selection remains sensitive to the kernel specification, whereas the $\Delta$-based decision rule provides more consistent results. Neither kernel specification consistently outperforms the other across all metrics.

\subsection{Sensitivity to the Threshold Parameter}
\label{app:exp2_threshold}

We further examine the sensitivity of the classification results to the threshold parameter $\epsilon$ in the experiment with observable parents. The AUC and Acc1 values are independent of $\epsilon$, while Acc2 and Acc3 depend on the threshold $n\epsilon$. The main experiments use $\epsilon=0.5$ for all settings without selecting different thresholds according to the underlying mechanism.

Table~\ref{tab:app_exp2_threshold} compares the results at $\epsilon=0.5$ with those obtained using empirically selected thresholds by grid search. For each kernel and mechanism, the threshold is selected by maximizing Acc3, with Acc2 used to resolve ties.

\begin{table}[t]
\caption{Sensitivity of Acc2 and Acc3 to the threshold parameter $\epsilon$ in the pairwise experiment with observable parents.}
\label{tab:app_exp2_threshold}
\centering
\begin{tabular}{llccc}
\toprule
Mechanism & Kernel & $\epsilon$ & Acc2 & Acc3 \\
\midrule
\multirow{2}{*}{Linear}
& \multirow{2}{*}{White} & 0.50 & 0.8889 & 0.8944 \\
& & 0.43 & 0.9000 & 0.9056 \\
\midrule
\multirow{2}{*}{Linear}
& \multirow{2}{*}{Amplitude} & 0.50 & 0.8944 & 0.9000 \\
& & 0.45 & 0.9000 & 0.9056 \\
\midrule
\multirow{2}{*}{Nonlinear}
& \multirow{2}{*}{White} & 0.50 & 0.9222 & 0.9500 \\
& & 3.00 & 0.9722 & 1.0000 \\
\midrule
\multirow{2}{*}{Nonlinear}
& \multirow{2}{*}{Amplitude} & 0.50 & 0.9333 & 0.9500 \\
& & 3.00 & 0.9833 & 1.0000 \\
\botrule
\end{tabular}
\end{table}

The empirically selected thresholds differ between linear and nonlinear mechanisms. For the linear cases, the preferred values are close to $\epsilon=0.5$, and the improvements over the fixed threshold are small. In contrast, the nonlinear cases favor larger thresholds, with $\epsilon=3$ for both kernels. These findings indicate that the preferred threshold depends on the causal mechanisms and that $\epsilon=0.5$ is not universally optimal. Nevertheless, the main experiments retain a single fixed threshold to avoid mechanism-specific threshold selection.

\section{Multi-node Sensitivity Analysis}
\label{app:exp3}

This section examines the sensitivity of PCG-CD to the decision order and kernel specification in the multi-node experiment. All comparisons use the same standardized datasets, a fixed threshold of $\epsilon=0.5$, and the same graph estimation procedure. We report Z-hit, E-hit, Sim1, and Sim2, following the definitions in the main text.

\subsection{Effect of Decision Order}
\label{app:exp3_decision_order}

We compare two decision orders for selecting causal relationships. The no-edge-first rule, adopted in PCG-CD algorithm, first checks whether the no-edge model has a shorter code-length than both directional models and then applies the $\Delta$-pseudo-collinearity criterion to the remaining pairs. In contrast, the $\Delta$-first rule applies the threshold $\Delta\leq n\epsilon$ before checking the no-edge model. The two rules are evaluated under the same kernel specifications and threshold.

Table~\ref{tab:app_exp3_order} presents the results under both decision orders. For both the White kernel and Amplitude kernel, the $\Delta$-first rule achieves higher Z-hit values in different sample sizes, exceeding the corresponding values obtained using the no-edge-first rule. However, the no-edge-first rule generally achieves higher E-hit and substantially higher Sim2.

These results indicate a trade-off between recovering true latent-confounded pairs and preserving the overall estimated causal structure. Applying the $\Delta$ threshold first assigns bidirected relations more readily when the directional code-lengths are similar according to Z-hit. The no-edge-first rule is more conservative in assigning potential latent-confounded relations because it first excludes pairs for which the no-edge model provides a shorter description. Neither decision order consistently achieves the highest values across all four metrics. We therefore retain the no-edge-first rule in the PCD-CD algorithm.

\begin{table}[t]
\caption{Effect of decision order on multi-node causal graph discovery under the White and Amplitude kernel specifications.}
\label{tab:app_exp3_order}
\centering
\begin{tabular}{llccccc}
\toprule
Kernel & Order & $n$ & Z-hit & E-hit & Sim1 & Sim2 \\
\midrule
\multirow{6}{*}{White}
& \multirow{3}{*}{No-edge-first}
& 200 & 0.5000 & 0.7717 & 0.9240 & 0.8360 \\
& & 500 & 0.3500 & 0.7333 & 0.9240 & 0.8600 \\
& & 1000 & 0.4000 & 0.7292 & 0.9100 & 0.8780 \\
\cmidrule(lr){2-7}
& \multirow{3}{*}{$\Delta$-first}
& 200 & 0.9000 & 0.6992 & 0.9180 & 0.5020 \\
& & 500 & 0.6500 & 0.7333 & 0.9160 & 0.4320 \\
& & 1000 & 0.8000 & 0.6667 & 0.9100 & 0.4660 \\
\midrule
\multirow{6}{*}{Amplitude}
& \multirow{3}{*}{No-edge-first}
& 200 & 0.6500 & 0.7525 & 0.9180 & 0.8340 \\
& & 500 & 0.3500 & 0.6917 & 0.9160 & 0.8480 \\
& & 1000 & 0.4500 & 0.6792 & 0.9060 & 0.8580 \\
\cmidrule(lr){2-7}
& \multirow{3}{*}{$\Delta$-first}
& 200 & 0.9500 & 0.7117 & 0.9220 & 0.4860 \\
& & 500 & 0.6500 & 0.7042 & 0.9220 & 0.4300 \\
& & 1000 & 0.8000 & 0.6375 & 0.9040 & 0.4600 \\
\botrule
\end{tabular}
\end{table}

\subsection{Effect of Kernel Specification}
\label{app:exp3_kernel}

We further compare the White and Amplitude kernel specifications while keeping the decision order and threshold fixed. Table~\ref{tab:app_exp3_order} also provides the results for this comparison.

Under the no-edge-first rule, the differences between the two kernel specifications are relatively small. Under the $\Delta$-first rule, the two kernels again produce similar results across the four metrics. Overall, the results indicate that the multi-node experiment is less sensitive to the kernel specification than to the decision order after data standardization. Neither kernel consistently outperforms the other across all sample sizes and metrics.

\end{appendices}


\bibliography{sn-bibliography}

\end{document}